\documentclass[a4paper,fleqn]{cas-dc}
\graphicspath{ {./figures/} }

\usepackage{amssymb}
\usepackage{multirow}
\usepackage{algorithmic}
\usepackage{array}
\usepackage{textcomp}
\usepackage{stfloats}
\usepackage{url}
\usepackage{verbatim}
\usepackage{graphicx}
\usepackage{epstopdf}
\usepackage{booktabs}
\usepackage{silence}
\usepackage{bm}
\usepackage[round,authoryear]{natbib}
\usepackage{subfigure}
\usepackage{amsmath,amsfonts}
\usepackage[ruled]{algorithm2e}
\usepackage{soul}
\usepackage{bm}
\usepackage{color, xcolor}

\soulregister \cite7
\soulregister \citep7
\soulregister\ref7

\def\tsc#1{\csdef{#1}{\textsc{\lowercase{#1}}\xspace}}
\tsc{WGM}
\tsc{QE}

\begin{document}
\let\WriteBookmarks\relax
\def\floatpagepagefraction{1}
\def\textpagefraction{.001}


\title [mode = title]{Occluded Person Re-Identification with Deep Learning: A Survey and Perspectives}



%
\author[1]{ Enhao Ning }
\fnref{fn1}


\ead{ ningenhao@163.com }
\author[1,2]{ Changshuo Wang }
\fnref{fn1}

\ead{ wangchangshuo@semi.ac.cn }

\fntext[fn1]{Equal contribution.}

\author[3]{Huang Zhang}
\ead{ zhhh1998@outlook.com }
\author[1]{ Xin Ning }
\cormark[1]
\ead{ ningxin@semi.ac.cn }
\author[4]{Prayag Tiwari }
\cormark[1]
\ead{prayag.tiwari@ieee.org}
\affiliation[1]{organization={Institute of Semiconductors, Chinese Academy of Sciences},
            city={Beijing},
            postcode={100083},
            country={China}}
\affiliation[2]{organization={Center of Materials Science and Optoelectronics Engineering $\&$ School of Microelectronics, University of Chinese Academy of Sciences},
            city={Beijing},
            postcode={100083},
            country={China}}
\affiliation[3]{organization={School of Software, Xinjiang University},
            city={Xinjiang},
            postcode={830000},
            country={China}}
\affiliation[4]{organization= {School of Information Technology, Halmstad University},
            city={Halmstad},
            postcode={30118}, 
            country={Sweden}}  


\cortext[1]{Corresponding author.}



\begin{abstract}
Person re-identification (Re-ID) technology plays an increasingly crucial role in intelligent surveillance systems. Widespread occlusion significantly impacts the performance of person Re-ID. Occluded person Re-ID refers to a pedestrian matching method that deals with challenges such as pedestrian information loss, noise interference, and perspective misalignment. It has garnered extensive attention from researchers. Over the past few years, several occlusion-solving person Re-ID methods have been proposed, tackling various sub-problems arising from occlusion. However, there is a lack of comprehensive studies that compare, summarize, and evaluate the potential of occluded person Re-ID methods in detail.
In this review, we start by providing a detailed overview of the datasets and evaluation scheme used for occluded person Re-ID. Next, we scientifically classify and analyze existing deep learning-based occluded person Re-ID methods from various perspectives, summarizing them concisely. Furthermore, we conduct a systematic comparison among these methods, identify the state-of-the-art approaches, and present an outlook on the future development of occluded person Re-ID.
\end{abstract}


\maketitle





\section{Introduction}
With the increasing integration and intelligence of surveillance equipment  \citep{bedagkar2014survey} in recent years, person re-identification (Re-ID) technology has significantly advanced. This technology finds extensive application in sensitive and specialized domains, such as medicine, rescue operations, criminal investigations, and surveillance. These fields often operate in complex and dynamic environments. Consequently, the rapid and accurate localization and identification of specific targets in multi-camera occlusion scenarios hold immense practical significance.

Given the complexity and variability of real-life scenes, where people and objects move randomly, and surveillance devices typically cover wide areas, the likelihood of occluded individuals is high. Occlusion can have a severe impact on visual information, rendering the affected features unreliable. Occlusion can occur due to object interference, changes in pedestrian pose, clothing, and perspective. 
In early pedestrian representations, researchers primarily relied on basic, local visual attributes extracted from images, such as color, texture, edges, and corner points. These features capture geometric shapes and pixel distributions in images but are highly sensitive to external factors, lacking robustness and generalization.
The development of deep learning has introduced high-level visual features. 
Compared with low-level visual features, high-level features are more adaptive to occlusions, noises and pose changes, and have stronger robustness in complex environments.
Consequently, numerous researchers have developed a multitude of methods to address the prevalent occlusion problem.  
In general, the occlusion problem is divided into three sub-problems: (1) Noise problem. The problem of interference by multiple and mixed information from the features in the acquisition of complex scenes. (2) Missing problem. The problem of incomplete pedestrian features is due to only a part of the pedestrian being captured. (3) Alignment problem. Owing to the change in posture, perspective, and position, the features cannot correspond one-to-one, which causes distraction, shared location misalignment, and other issues.
The study of occlusion also involves the separation of humans and backgrounds to extract human features as the core. Methods to extract fine-grained, highly discriminative, and more essential features with reference and value have also been studied.

\begin{figure*}[t]
	\centering
	\includegraphics[scale=0.35]{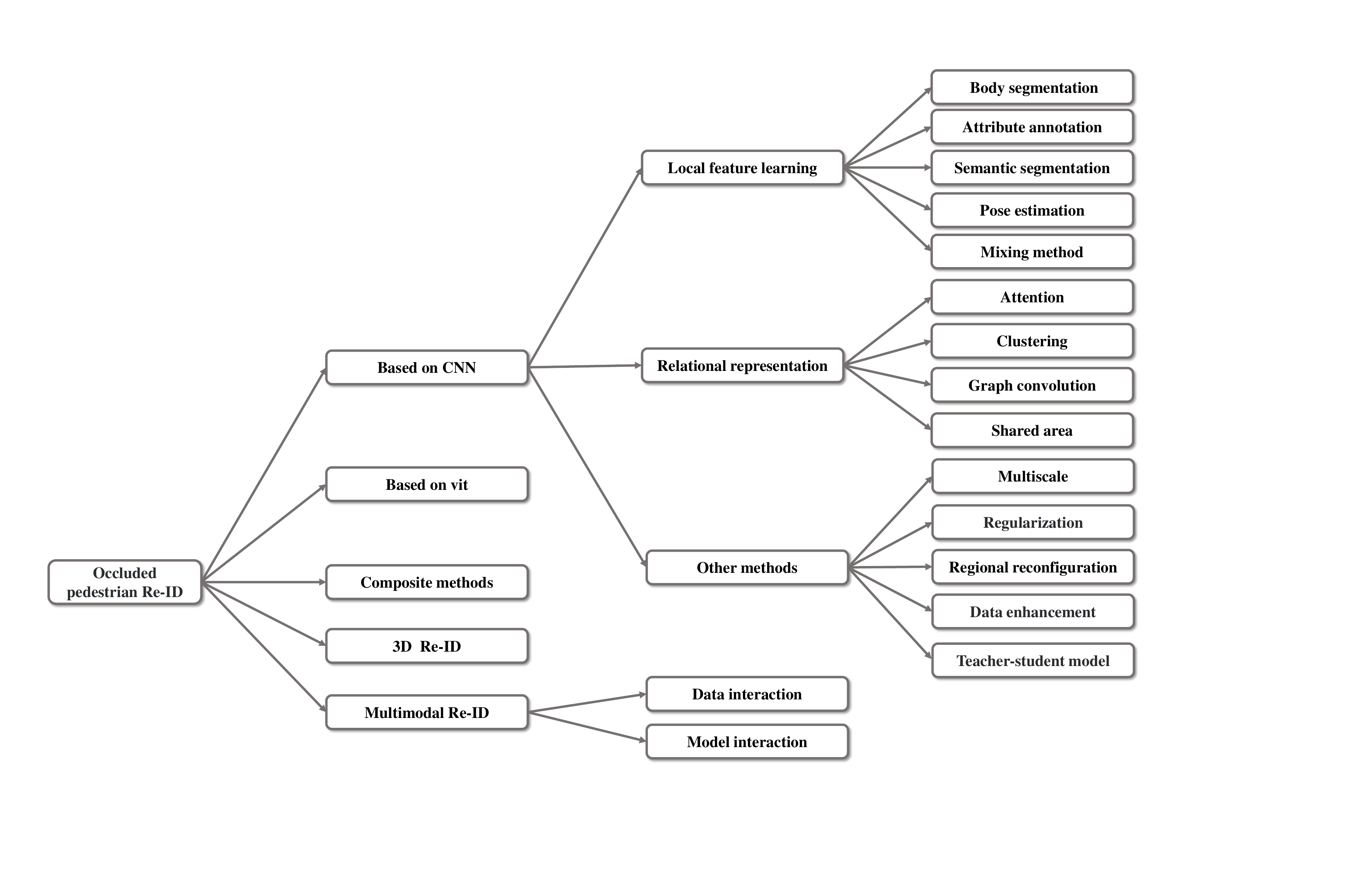}
	\caption{{Overall structure of the survey.}  }
       \label{t2}
\end{figure*}

We want to identify the current state-of-the-art and limitations of existing methods and discover unexplored areas. Specifically, we present methods for dealing with occluded person Re-ID that were submitted in top international journals or conferences before 2023. We classify deep learning-based occluded person Re-ID according to the network structure of extracted features (CNN-based, transformer-based, and hybrid structure-based), the way features are extracted (uni-modal and multi-modal), and the hierarchical structure of features (2d and 3d).  ( see Figure \ref{t2}). 
First, due to the powerful performance of convolutional neural networks (CNNs) in image matching tasks, CNN-based methods have become one of the mainstream methods to deal with occlusion problems in person Re-ID. Therefore, we treat the cnn-based methods as the first class of methods to deal with the occlusion problem.
Secondly, based on the success of transformer in the field of natural language, in recent years, vit has also been widely used to deal with the occlusion problem in pedestrian re-identification with good results. Therefore, we treat transformer-based methods as the second category.
The third class of methods are some composite methods. For example, the complementary nature of CNN and vit is exploited to form a hybrid structure. 
The fourth and fifth class of methods are based on 3D and multimodal to deal with the occlusion problem in person Re-ID. They deal with more scenarios and are a relatively novel approach.

\begin{table*}[t]
\caption{\label{table1} The summary of person-reported surveys in recent years.}
\centering
\resizebox{170mm}{25mm}{
\begin{tabular}{c|c} 
\hline
Survey                                                                                              & Venue      \\ 
\hline
A survey of approaches and trends in person Re-ID \citep{bedagkar2014survey}                                    & IVC2014    \\
Person Re-ID Past, Present and Future \citep{zheng2016person}                                              & arXiv2016  \\
A systematic evaluation and benchmark for person Re-ID: Features, metrics, and datasets \citep{gou2018systematic} & TPAMI2019  \\
Beyond intra-modality discrepancy: A comprehensive survey of heterogeneous person Re-ID \citep{wang2019beyond} & arXiv2019  \\
A Survey of Open-World Person ReIdentification \citep{leng2019survey}                                                      & TCSVT2020  \\
Survey on Reliable Deep Learning-Based person Re-ID Models: Are We There Yet? \citep{lavi2020survey}           & arXiv2020  \\
Deep Learning for Person Reidentification: A Survey and Outlook \citep{ye2021deep}                                     & TPAMI2021  \\
SSS-PR: A short survey of surveys in person Re-ID \citep{yaghoubi2021sss}                                       & PRL2021    \\
Deep learning-based person Re-ID methods: A survey and outlook of recent works \citep{ming2022deep}          & IVC2022    \\
Deep Learning-based Occluded person Re-ID: A Survey \citep{peng2022deep}                                     & arXiv2022  \\
\hline
\end{tabular}}
\end{table*}
In general, the contributions of this study are as follows:

1) This study focuses on addressing the occlusion problem in person Re-ID models, which is crucial for achieving high accuracy and robustness. We present a scientific and comprehensive review of past and current state-of-the-art approaches.

2) The current review of person Re-ID methods lacks sufficient coverage of approaches based on ViT. Given the excellent performance of ViT in occluded person Re-ID, we include a discussion of this method and its hybrid variants in our study, offering researchers new ideas and options for addressing the occlusion problem.

3) We creatively incorporate 3D person Re-ID and multimodal person Re-ID, which have become popular in recent years. These novel methods can better solve the occlusion problem by utilizing additional depth or modal information, thus improving the performance and reliability of person Re-ID.

4) We anticipate advancements in occluded person Re-ID and firmly believe that continued research and innovation will lead to the development of more effective methods and technologies for addressing the occlusion problem. These advancements will inspire and drive progress in the field of person Re-ID.

\section{ Literature review }
In the field of person Re-ID, there is a relative scarcity of specialized reviews compared to methodological articles.
And they all focus on specific problems. These surveys are listed in Table \ref{table1}. 
 \cite{bedagkar2014survey} focuses on the challenges of person Re-ID and divides it into open-set Re-ID and closed-set Re-ID based on the fixity of the gallery. 
 \cite{zheng2016person} divides the methods of person Re-ID into methods for images and methods for videos based on the matching strategy. 
 \cite{gou2018systematic}  provides a more detailed study of the features, metrics, and datasets of person Re-ID.
 \cite{wang2019beyond} focuses on heterogeneous person Re-ID. According to the application scenario, it classify the methods into four categories — low-resolution, infrared,sketch, and text. 
 \cite{leng2019survey} focuses on open-world Re-ID tasks.
 \cite{lavi2020survey} classifies Re-ID into single feature learning based approaches and multi-feature learning based approaches based on feature learning strategies.
 \cite{ye2021deep} provides a more detailed explanation of Re-ID for open and closed settings, and introduces methods such as transmembrane states, unsupervised.
 \cite{yaghoubi2021sss} provides a more multidimensional classification of the person Re-ID problem.
 \cite{ming2022deep}  classifies the methods of person Re-ID into four categories based on metric learning and representation learning, and adds the latest methods.
 \cite{peng2022deep} focuses on image-based obscured person Re-ID methods. However, these investigations are inevitably affected by a number of inherent limitations.
Considering the widespread existence of the occlusion problem in pedestrian recognition, research on occluded person Re-ID is essential. Therefore, we provide an in-depth summary and comprehensive analysis of methods and prospects in occluded person Re-ID to advance future developments.

\section{ Datasets and Evaluation Protocols }

\subsection{ \textbf{Datasets} }
\begin{figure}[!t]
	\centering
	\includegraphics[scale=0.40]{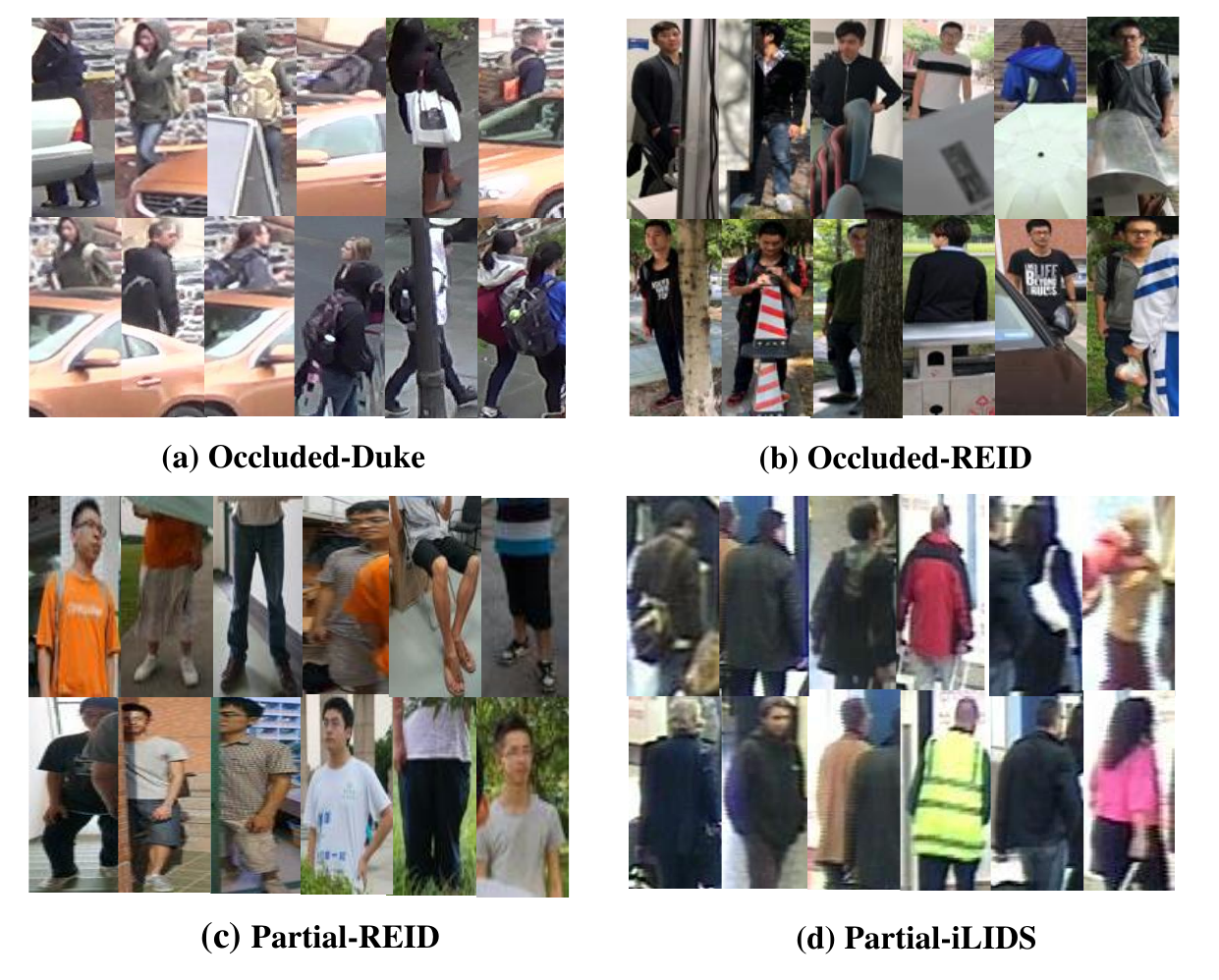}
	\caption{ Examples of four commonly used occluded person Re-ID datasets.  }
       \label{t1}
\end{figure}

Occluded person Re-ID datasets can be divided into two categories: partial person and occluded person Re-ID datasets. The pedestrian images of the occluded person Re-ID datasets have occlusion information interference and are not cropped. The pedestrian image portion of the partial person Re-ID dataset is present and artificially cropped.
Examples of partial/occluded person Re-ID datasets are shown in Figure \ref{t1}.

\textbf{Occluded-DukeMTMC } \citep{miao2019pose} was collected from DukeMTMC-reID \citep{zheng2017unlabeled}, containing 15,618 training images of 708 pedestrians, 2,210 query images of 519 pedestrians, and 17,661 gallery images of 1,110  pedestrians for testing. Of these images, 9$\%$ of the training set, 100$\%$ of the query set, and 10$\%$ of the gallery are occluded images. Obstacles include cars, bicycles, trees, and other pedestrians, adding complexity to the dataset.

\textbf{P-ETHZ  } \citep{zheng2015partial} is an image-based occluded person Re-ID dataset, modified by ETHZ  \citep{ess2008mobile}. It has 3,897 images containing 85 pedestrian identities with 1 to 30 full-body and occluded pedestrian images per identity.

\textbf{P-DukeMTMC-reID  } \citep{zhuo2018occluded} was modified from DukeMTMC-reID  \citep{zheng2017unlabeled}, containing a total of 24,143 images of 1,299 pedestrians, and each identity has a full-body and occlusion image; the pedestrian in the image is occluded by different objects, such as other pedestrians, cars, and signage.

\textbf{Occluded-REID } \citep{zheng2015partial} has 2,000 images of 200 pedestrians, each pedestrian corresponding to 5 occlusion and 5 whole body images, collected from Sun Yat-sen University. 
The dataset includes different viewpoints and types of severe occlusion, which challenges person Re-ID.

\textbf{Occluded-DukeMTMC-VideoReID 
  } \citep{hou2021feature} was reorganized from the DukeMTMC-VideoReID  \citep{wu2018exploit} dataset. The training set contains 1,702 trajectory segments covering 702 pedestrians, the test set queries cover 661 pedestrians, and the gallery covers 1,110. More than 70$\%$ of the videos are occluded, including different perspectives and a variety of obstacles, such as cars, trees, bicycles, and other pedestrians.

\textbf{Partial-ReID  } \citep{zheng2015partial} has 600 images of 60 pedestrians, 5 partial and 5 full-body images for each pedestrian.  Using the visible parts, they are manually cropped to form new partial images. 
The images are collected from different perspectives, backgrounds and occlusions in a university campus.

\textbf{Partial-iLIDS  } \citep{he2018deep} was derived from iLIDS   \citep{zheng2011person} and contains 238 images of 119 pedestrians. Each pedestrian corresponds to one manually cropped non-occluded partial image and one full-body image. The partial image is used as a query, and the full-body image is used as a search library. It was shot by multiple non-overlapping cameras, mostly for test sets.

\textbf{Partial-CAVIAR  }\citep{he2018deep} was derived from CAVIAR  \citep{cheng2011custom} and contains 142 images of 72 pedestrians.  The partial map is generated by randomly picking half of the overall image of each pedestrian.

\textbf{P-CUHK03  } \citep{kim2017deep} was constructed based on CUHK03  \citep{li2014deepreid}, with a total of 1,360 pedestrian images, wherein 15,080 images corresponding to 1,160 pedestrians are used as a training set, and the remaining 100 pedestrians are used as a validation and test set.  
Two of the images are selected to generate 10 local body query images with a spatial area ratio, and the remaining three images are used as whole body gallery images. 

\subsection{\textbf{ Evaluation Protocols} }
In the field of occluded person Re-ID, the commonly used evaluation metrics are Cumulative Matching Characteristic (CMC) curves and mean Average Precision (mAP).

CMC curves are based on the principle of ranking the similarity between the query image and  the image library, and the higher the top image,  the higher the similarity with the query image.
Then, the top-k accuracy $ {\rm \emph{ACC}}^\emph{k}$ of the query image is calculated based on this ranking. 
If the first k samples contain the query target, then $ {\rm \emph{ACC}}^\emph{k}$ is 1, $\emph{k}\in\{1,2,3...\}$. Otherwise, $ {\rm \emph{ACC}}^\emph{k}$ is 0 .
Finally, the $ {\rm \emph{ACC}}^\emph{k}$curves for all targets are summed and divided by the total number of targets to obtain CMC-k.

mAP better reflects the degree to which all correct target pictures are at the top of the sorted list. Compared with the CMC curve, it can more comprehensively measure the performance of Re-ID algorithms, where P is the precision rate, which refers to the proportion of correct samples among all samples. It reflects the accuracy of the correct samples in the output. The AP is the average of all correct samples predicted by the model. It reflects how well the model works on a single category and is the average of the accuracy of each correct prediction. Since there is more than one class in the recognition, the average AP value needs to be calculated for all classes, so the average accuracy of each class is added and divided by the total number of classes to obtain mAP.

The CMC curve cannot consider the hits of the samples with lower rankings, while mAP takes all samples into account. Therefore, they are important and complementary.

\section{ Deep Learning Methods }
\subsection{ \textbf{Based on CNN } }
Convolutional neural networks (CNNs) have emerged as one of the leading methods for learning pedestrian representations from RGB images. By using local perceptual fields and learning filters, CNNs can extract powerful features that capture regional information about local features of pedestrians. These features are then compressed and mapped to higher-level representations. Researchers have refined them to be usable for pedestrian matching tasks in complex realistic scenarios.
We classify it into local feature learning, relational representation,  mixing methods, and other methods.

\subsubsection{ \textbf{Local Feature Learning }}
\begin{figure}[t]
	\centering
	\includegraphics[scale=0.50]{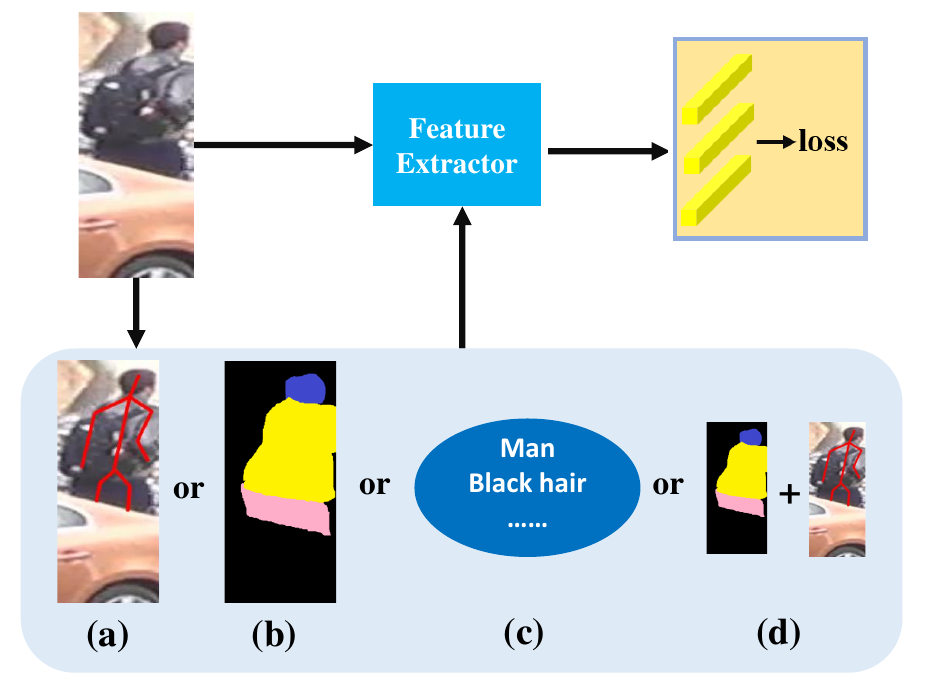}
	\caption{  Four different local feature learning methods: (a) indicates pose estimation. (b) indicates semantic segmentation. (c) indicates attribute annotations. (d) indicates the mixing method. }
       \label{t3}
\end{figure}

Local feature-learning performs better at handling regional features, and it has unique advantages for occlusion region recognition and location compared with global features. According to its implementation of different local feature methods, we divide them into human segmentation, pose estimation, human parsing, attribute annotation, and hybrid methods ( see Figure \ref{t3} ).

\begin{figure}[t]
	\centering
	\includegraphics[scale=0.42]{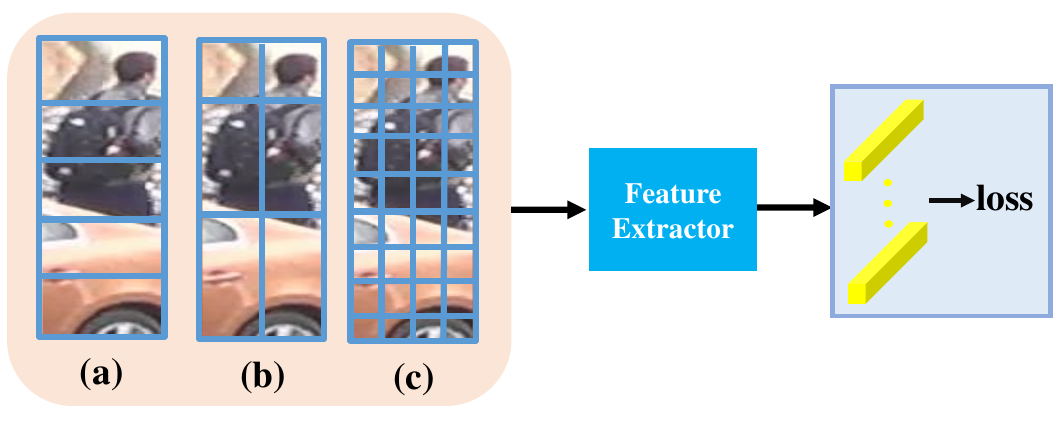}
	\caption{  Three common Body Segmentation schematics: (a) indicates stripes. (b) indicates fixed areas. (c) indicates small blocks.  }
       \label{t4}
\end{figure}

\textbf{Body Segmentation.} By leveraging the characteristic of pedestrians walking upright, our method extracts improved local features through the segmentation of the original image or feature map. The segmentation results can take the form of stripes, fixed regions, or small patches ( see Figure \ref{t4} ).

However, segmentation does not have the process of identifying occlusions, so it is sensitive to noise. 
CBDB-Net  \citep{tan2021incomplete} evenly divides the strips on the feature map and discards each strip one by one to output multiple incomplete feature maps, which forces the model to learn a more robust pedestrian representation in an environment with incomplete information.  
The DPPR  \citep{kim2017deep} predefines thirteen bounding boxes for the whole-body image, including the whole image, half-body image, and horizontal part image, and extracts features from each part. 
At the same time, an attention-based matching mechanism is introduced to make the feature weight of the same body part larger, which can alleviate the information loss caused by occlusion.
At the same time, an attention-based matching mechanism is introduced to make the feature weight of the same body part larger, which can alleviate the information loss caused by occlusion. 
OCNet  \citep{kim2022occluded} introduces a relationship-based approach to deal with occlusion problems.  
OCNet  \citep{kim2022occluded} divides the feature map horizontally into top and bottom features and takes 1/4 of the middle width as the central feature. Then, it is put into the relational adaptive module consisting of two shared layers together with the global feature map. The alignment problem between regional features is handled by relations, and weights are introduced to suppress noise interference.

\textbf{Pose Estimation.}  Pose estimation extracts semantic information at the image pose level by exploiting the highly structured human skeleton. The interference of noise is suppressed in a guided or fused manner.

HOReID  \citep{wang2020high} introduced a learnable relational matrix.  The human body key points obtained from the pose estimation are regarded as nodes in the graph, and finally, a topology graph is formed to suppress noise interference.  
PMFB  \citep{miao2021identifying} uses pose estimation to obtain confidence and coordinates of human keypoints.  
Then, a threshold is set to filter the occluded regions.  
Finally, the visible part is used to constrain the feature response at channel level to solve the occlusion problem. 
 PGMANet  \citep{zhai2021pgmanet} generates an attention mask using a human heat map.  
The interference of noise is removed jointly by the dot product of feature maps and guidance of higher-order relations.

Researchers generally use pose estimation in two directions: 1) to obtain semantic features through pose estimation, identify noise points, and better remove noise interference, and 2) to localize human regions through pose estimation and thus solve the problem
of alignment and local feature extraction. AACN  \citep{xu2018attention}  uses pose points to locate pedestrian body regions and introduces a posture-guided visibility score to separate occlusions.
 DAReID  \citep{xu2021dual} adopts a dual-branch structure, in which the mask branch extracts more discriminative local features based on the spatial attention module guided by pose estimation, and the global branch enhances the representation of human discriminative information through feature activation.  
DSA-reID  \citep{zhang2019densely} estimates dense semantic information at the 2D level by a pre-trained DensePose  \citep{guler2018densepose} model and maps a set of dense 3D semantic alignment components. 
 Local features are extracted by integrating neighboring components. 
Finally, the global and local features are fused into the final feature.  
PGFL-KD  \citep{zheng2021pose} takes the local features of the semantic layer as queries and looks for more prominent foreground regions in the feature map to obtain enhanced foreground features.  
Based on this feature, interactive training and knowledge distillation are performed to constrain the learning of the backbone network.  
ACSAP  \citep{he2021adversarial} uses pose keypoints to guide the adversarial generation network to remove noise interference by weakening the spatial relationship between the front and back blocks.
PDC  \citep{su2017pose} obtains 6 body regions according to 14 key points of pose, rotates and scales each part, and uses an improved PTN network to learn the parameters of affine transformation. 
These local features are automatically placed at certain locations in the drawing to resolve alignment issues. 
PVPM  \citep{gao2020pose} trains a visibility predictor based on the correspondence between visibility parts. After that, the alignment problem is solved by generating part pseudo-labels through graph matching.

\textbf{Semantic Segmentation.} By introducing a human parsing model, the interference of noise is identified and removed in the form of segmentation or semantic parsing.

 SPReID  \citep{kalayeh2018human} generates probability maps associated with five different body regions based on the trained pedestrian class semantic parsing model Inception-V3  \citep{szegedy2016rethinking}, namely, foreground, head, upper body, lower body, and shoes. Then, the probability map is fused with the semantic region features after bilinear interpolation to activate different parts and remove the interference of occlusion. CoAttention  \citep{lin2021self} takes the parsing mask of the local image of the pedestrian’s body as a query and constructs a mapping while introducing a self-attentive mechanism to filter occlusions.
MMGA  \citep{cai2019multi} first separates pedestrians from images into upper and lower body using JPPNet  \citep{liang2018look}. 
Then, two attention modules are designed; the first is used to filter the interference from the background, and the second generates the corresponding spatial and channel attentions to extract different features guided by the whole, upper, and lower body masks. Finally, element-level multiplication is performed as the final feature.
HPNet  \citep{huang2020human} uses the COCO  \citep{lin2014microsoft} dataset to train the human body parsing model to obtain labels for the four main body parts, based on which the parsing model and  overall network are trained in a multitasking manner while generating visibility scores to remove occlusions.

Semantic segmentation-based approaches also contribute to enhancing the diversity of features. In the case of SORN \citep{zhang2020semantic}, a three-branch model composed of a global branch, a local branch, and a semantic branch is designed. The global branch aims to obtain global features through normalization and feature aggregation. Meanwhile, the local branch leverages prior knowledge of the pedestrian body structure to generate pedestrian body parts and derive local features through mapping, pooling, and normalization. The semantic branch first uses the DANet  \citep{fu2019dual} model to pre-train the semantic labels of the data, trains a semantic segmentation model on the DensePose-COCO dataset   \citep{guler2018densepose}, introduces label smoothing to optimize the semantic labels, and forms a foreground pedestrian body region by aggregating the semantic segmentation part to realize the separation of background and pedestrians. SGSFA  \citep{ren2020semantic} introduces a semantic alignment branch and  spatial feature alignment branch.  The former achieves semantic alignment through element-level multiplication.  The latter is based on regional spatial alignment achieved by body structure.

\textbf{Attribute annotation.} The occlusion problem is handled by introducing attribute annotation.

ASAN  \citep{jin2021occlusion} extracts the visible part of human features by combining attribute information and weak supervision. Attribute information is a semantic level attribute annotation. 
Based on the visibility part determination, a region visibility matching algorithm is introduced to achieve the effect of denoising.

\textbf{Mixing method.} Introducing more than two kinds of external information can help the model remove the interference of noise in the form of feature interaction or co-guidance.

GASM  \citep{he2020guided} proposes an architecture for learning salient information, which separates pedestrians from background by using semantic information, removes occlusion interference by pose estimation, and then fuses the two features to guide model learning.  
SSPReID  \citep{quispe2019improved} designs a joint learning method to combine salient and semantic features. Five different semantic features of the human body are obtained by human body parsing. The saliency features utilize the regions of highest attention in the graph. These features are fused with global features and finally concatenated together to form the final features. 
TSA  \citep{gao2020texture} introduces two kinds of area features for the pose change problem, one guided by pose keypoints and one guided by partial masks of the human parsing model, after which the two are fused. 
Using interaction can solve the pose change problem, while using pose and segmentation can also suppress noise and thus solve the occlusion problem.
FGSA  \citep{zhou2020fine} proposes a pose resolution network for complex pose changes to deal with local locations and the relationships between them. Attribute interaction learning is designed for local feature information extraction corresponding to pose points, which is achieved by training an intermediate attribute classification model that treats attribute recognition as a multi-category labeling problem. Finally, a local enhanced alignment model is added in the feature fusion phase, that is, less weight is added to the background and more weight is added to the local and attribute locations. The backbone network of LKWS  \citep{yang2021learning} is based on PCB  \citep{sun2018beyond}. In local feature extraction, the visibility label of points is generated by pose estimation and a reasonable threshold, and then the visibility of thick stripes is obtained by a voting mechanism. Based on the visibility of stripes, a visibility discriminator is trained to recognize noise interference.  
PGFA  \citep{miao2019pose} introduces a two-branch structure, where the local branch extracts local features based on horizontal segmentation. 
The global branch is guided by pose, and key points are first obtained from pose estimation. 
Then, a reasonable threshold is set to filter the noise points, a proper dot product is performed with the feature map to fuse the features, and the features of the partial branches are stitched together into the final global features.
Similarly, PDVM  \citep{zhou2020depth} extracts the de-obscured global features based on the pose heat map. Segmentation guides local feature extraction.

 \begin{figure}[t]
	\centering
	\includegraphics[scale=0.40]{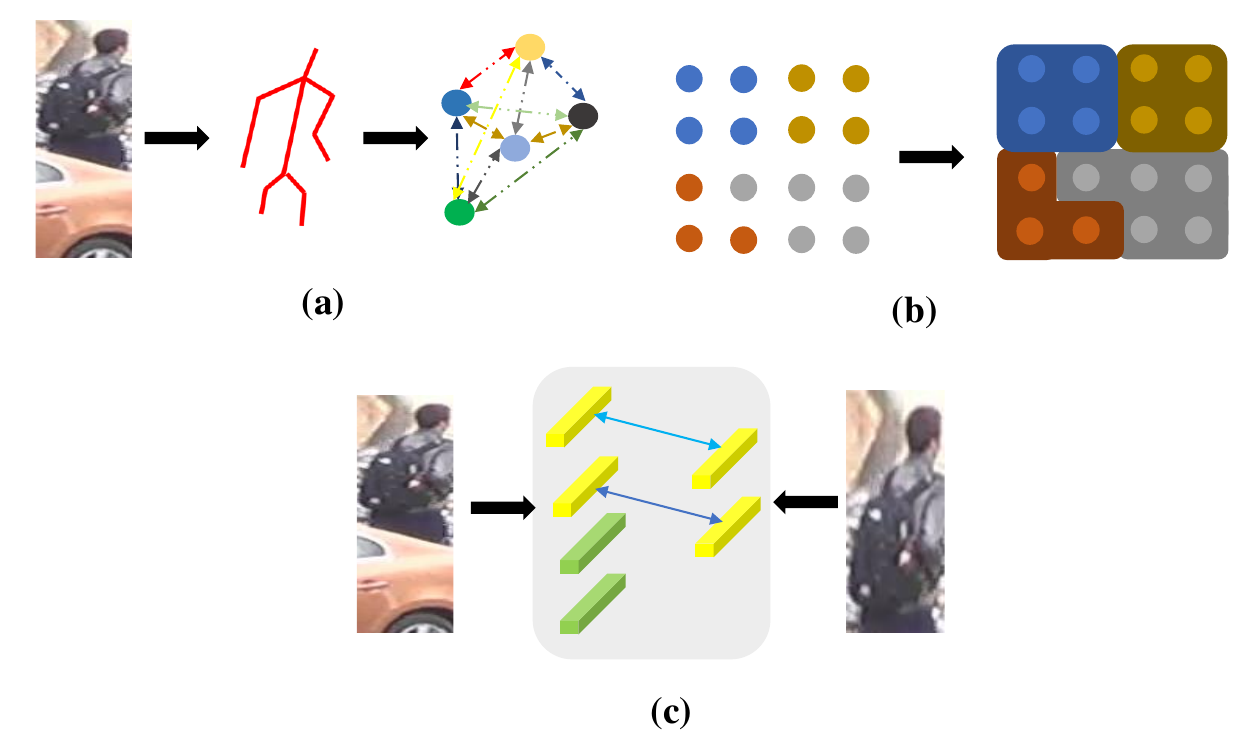}
	\caption{ (a) Schematic diagram showing figure convolution. (b) Schematic representation of clustering. (c) Schematic representation  ofrepresentation of the shared area.   }
       \label{t5}
\end{figure}
\subsubsection{  \textbf{Relationship Representation }}
By focusing on feature relationships, occlusions are handled in a suppressed, removed, or supervised manner.  We divide them into attention, clustering, graph convolution, and shared region ( see Figure \ref{t5} ) according to their different ways and means of learning relationships

\textbf{Attention.}  By introducing the attention mechanism, the model can select the highly salient and discriminative regions to suppress the interference of noise.

To address the issue of missing images in person Re-ID, DPPR  \citep{kim2017deep} employs an attention mechanism to emphasize the same pedestrian part across different images. This approach enhances the representation of individuals and improves matching accuracy.
Moreover, OCNet  \citep{kim2022occluded} mitigates the effect of noise by capturing higher-order relationships among regional features and incorporating them with weighted combinations. This method effectively suppresses the influence of noisy or irrelevant information, resulting in more robust and accurate person Re-ID outcomes. 
AACN  \citep{xu2018attention} combines pose guided attention maps and partial visibility scores to remove background distractions and occlusions before extracting clean pedestrian features.
DAReID  \citep{xu2021dual} introduces dual attention recognition. The local area visible to the pedestrian is obtained by gesture-guided spatial attention.  Global features are extracted by feature activation and pose. Both will then be used together to guide the representation of features. 
PISNet  \citep{zhao2020not} is concerned with the overlapping area between people and objects. A module is designed to act as a guide feature by querying features, which can attenuate the problem of attention distraction caused by multiple pedestrians in the gallery. Also, a reverse attention module based on strong activation attention is designed, which enables the model to assign more weight to the target region. 
APN  \citep{huo2021attentive} proposes a partial perceptual attention network, which takes partial feature maps as query vectors, calculates a similarity mapping M with a mapping X of the feature maps, and morphs the features by weighting M by X to achieve the purpose of aggregating and extracting refined features. 
MHSA-Net  \citep{tan2022mhsa} multiplies attention weights with feature maps and applies a nonlinear transformation to encourage multi-headed attention mechanisms to adaptively capture key local features. CASN  \citep{zheng2019re} takes attention and attention consistency as the criteria for model learning and removes occlusion by introducing an attention twin network to focus on more discriminative core areas. 
VPM  \citep{sun2019perceive} learns the visibility and location of components by introducing a self-supervised component localizer at the convolution output and introducing a feature extractor that generates region information through weighted pooling.
PSE  \citep{sarfraz2018pose} solves the occlusion problem caused by view angle using view angle prediction.

Attention-based approaches not only enhance the flexibility of the model, but also inject more contextual information into the features. 
PAFM  \citep{yang2022pafm}  introduces an improved spatial attention module to discover relationships between pixel points while capturing and aggregating pixel points with high semantic relevance. Finally, it is multiplied with the feature map containing pose information to perform feature fusion. 
Co-Attention  \citep{lin2021self} takes the analytic mask of a partial pedestrian image and whole image as the target and matches it through a self-attention mechanism  \citep{li2020effective}. 
Finally, noise interference is suppressed by focusing pedestrian features. 
QPM  \citep{wang2022quality} divides the feature map into six parts evenly in the vertical direction, and introduces a component quality score to judge the visibility. 
Meanwhile, a two-layer identity-aware module based on an attention map is used to deal with pedestrian occlusion in the global branch.  Finally, global features are adaptively extracted from clean pedestrian regions. 
DSOP  \citep{wang2020deep} divides occlusion into shallow and deep layers.  The shallow layer learns the feature after occlusion by focusing on the local region, and the deep layer gives a large receptive field to learn the global feature, that is, the feature before occlusion. After that, the channel and spatial attention mechanisms are applied to the two branches for weighted fusion of features  \citep{ning2021jwsaa}.

\textbf{Clustering.}The interference of noise is solved by finding the inherent distribution structure of the data to categorize the pixel points.

ISP  \citep{zhu2020identity} assigns a pseudo-label to each pixel by tandem clustering. All pixels of the human body image are firstly divided into foreground and background, based on the assumption that the foreground is more responsive than the background. Secondly, the pixels are clustered into different parts and assigned pseudo-labels. Based on the pseudo-labels, different weights are assigned to the pixels to extract local features. This not only separates occlusions from pedestrians at the pixel level, but also enables automatic alignment.

\textbf{Graph Convolution.}
By learning the high-order semantic relationship between pixels, the noise interference is suppressed by restricting the information transmission.

HOReID  \citep{wang2020high} introduces a matrix describing the higher order relationships between points and later passes information in this relationship matrix to form a topological map. With the help of the constraints of the topological map, the transfer of useless information between points is suppressed, and the purpose of noise removal is achieved.

\textbf{Shared Area.} The interference of noise is reduced by sensing the same body parts of pedestrians in the image pairs to extract shareable features.

DPPR  \citep{kim2017deep} gives larger weights to regions containing the same body parts to improve the ability of the model to extract core features.
VPM  \citep{sun2019perceive} solves the alignment and denoising problem by perceiving the visibility of shared regions.
PPCL  \citep{he2021partial} learns component matching in a self-supervised manner and finally computes image similarity based on shared semantic corresponding regions only. 
KBFM  \citep{han2020keypoint} focuses on extracting highly visible and shareable pose points, which are used as the core area to extract features for matching, and achieve the effect of denoising and alignment.

\subsubsection{  \textbf{Other Methods}}
\textbf{Regional reconfiguration.}This is complements obscured or noisy areas by using complete pedestrian areas.

To solve the problem of information loss caused by occlusion,  RFCNet  \citep{hou2021feature} introduces an encoder–decoder that uses non-occlusion remote spatial context for feature completion. The encoder is modeled by similarity region assignment. The decoder reconstructs the occluded region by establishing the correlation between the occluded region and distant non-occluded region through clustering.
ACSAP  \citep{he2021adversarial} combines attitude and adversarial generation networks and designs an attitude-guided spatial generator and spatial discriminator to remove noise interference.

\textbf{Data enhancement.} The sensitivity of the model to occlusion is improved by incorporating transformation of the data.

APNet  \citep{zhong2020robust} proposes a method to modify the detected bounding box. 
APNet  \citep{zhong2020robust} designed a bounding box aligner that slides over the image in a matching manner. Then, a feature extractor with high discriminative power is designed to extract the core local features and discard the noisy local features. IGOAS  \citep{zhao2021incremental} adopts a progressive occlusion module, which randomly generates small uniform occlusion on a group of images and generates larger occlusion based on small occlusion after model learning. Such a growing occlusion region can improve the recognition ability of occlusion and achieve the purpose of removing occlusion noise. OAMN  \citep{chen2021occlude} adopts a method based on cropping and scaling, predefines four corners, randomly selects a training image to be cropped and scaled to form patches on four positions, and realizes weighted learning in combination with attention to achieve denoising. RE  \citep{zhong2020random} introduces the technique of random pixel removal, which replaces the pixel values in the region with random values by randomly selecting rectangular regions, thereby improving the diversity of data and robustness of the model. SSGR  \citep{yan2021occluded} introduces a compound batch erase method, which includes two erase operations: one is frequently used random erase, and the other is batch constant erase. It first divides the image horizontally into random S and randomly selects a strip in each sub-batch to erase. Then, referring to the self-attention mechanism and local feature learning, a matching-based disentanglement non-local operation is introduced to extract better features from the complete pedestrian region. 
ETNDNet \citep{dong2023erasing} addresses the occlusion problem from an adversarial defence perspective. It deals with incomplete information, positional misalignment and noisy information through strategies of randomly erasing feature maps, introducing random transformations and perturbing feature maps.

\textbf{Regularization.} By imposing penalties and constraints on high-attention areas, the model is forced to focus on the full pedestrian area. Pedestrian features are extracted using complete information.

MHSA-Net  \citep{tan2022mhsa} introduces a feature regularization mechanism, which consists of a regularization term based on attention weight embedding and a hard triplet loss based on triplet feature units. The regularization term can cover local information in many ways and increase the completeness of information  \citep{ning2020feature}. Hard triplet loss can refine the fusion and be better used for pedestrian matching.

\textbf{Teacher-student model.} Teacher models assist student models in dealing with occlusion problems.

HG  \citep{kiran2021holistic} designs an end-to-end unsupervised teacher–student framework that lets the teacher network learn the between-class distance by inputting different combinations of images, and then the student inherits the network and learns the within-class distance by inputting more noisy images of the same class. 
At the same time, the attention embedding method with distance distribution matching can help the student network to remove noise interference better and extract more discriminative features. 
AFPB  \citep{zhuo2019novel} first puts regular data and pedestrian volume data simulating occlusion into the teacher network, and then it conducts joint training to make the teacher network learn a basic model that is robust to occlusion. The student network then inherits the teacher network. It learns on more realistic, noisier real-world occluded pedestrian
data.

\textbf{Multiscale.} The occlusion problem is handled by multi-scale feature representation.

DSR  \citep{he2018deep} focuses on solving problems caused by differences in scale. 
DSR \citep{he2018deep} first trains a fixed-size fully convolutional network with only convolution and pooling on Market-1501  \citep{zheng2015scalable} to represent the identity feature map, and it then introduces three different scales of blocks to extract features in a sliding manner \citep{li2021person}. 
Similarly, FPR  \citep{he2019foreground} also introduces a structure that only has convolution and pooling and proposes a pyramid layer composed of multiple different pooling kernels to solve the scale problem, an attention-based foreground probability generator to process the background, and a small weight to the background to achieve removal of background distractions.

\begin{figure}[t]
	\centering
	\includegraphics[scale=0.40]{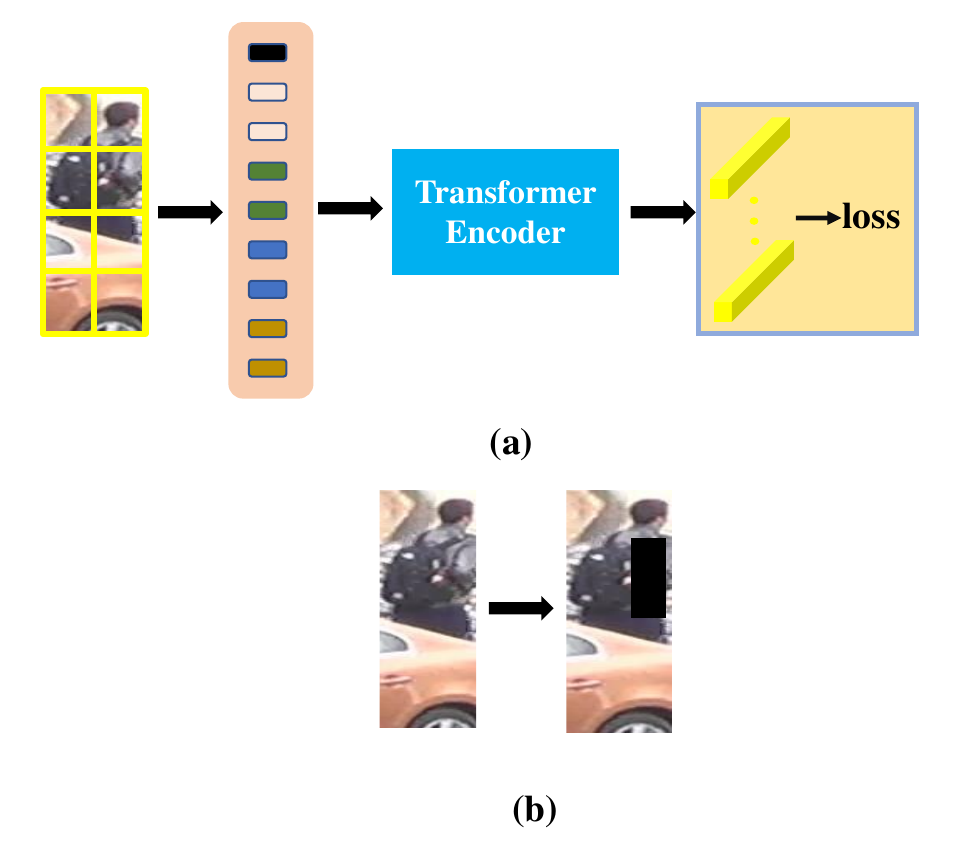}
	\caption{ (a) Schematic of the transformer-based approach. (b) Schematic representation of data augmentation.
 }
       \label{t6}
\end{figure}
\subsection{ \textbf{Based on vit } }
TransReID  \citep{he2021transreid}, proposed in 2021, is the first model to apply a vit  \citep{dosovitskiy2020image} in Re-ID ( see Figure \ref{t6} ), which has two major features compared to ResNet  \citep{he2016deep}: (1) Multi-headed self-attention is better at capturing long-range relationships and driving the model to focus on different body  parts  \citep{khan2022transformers, shamshad2023transformers}. (2) The transformer can retain more detailed information in the absence of downsampling computation. 
Based on these characteristics, many variants of transformer have appeared in recent years, and researchers have widely used them in occlusion Re-ID.

PFD  \citep{wang2022pose} first divides the image into fixed-size blocks that can be overlapped and then uses the transformer encoder to capture contextual relationships. Next, pose guided feature aggregation and a feature matching mechanism are used to display the visible body parts. Finally, the pose heat map and decoder are used as keys and values to learn a set of semantic part views to enhance the discriminability of body parts. 
DPM  \citep{tan2022dynamic} introduces subspace selection to deal with the alignment problem.  Specifically, DPM  \citep{tan2022dynamic} first aggregates the feature representations of the overall prototype map and the occlusion map using hierarchical semantic information to enhance the quality of the generated prototype mask.  Then, a head enrichment module based on normalization and orthogonality constraints is introduced to enhance the discriminative representation of features and remove the interference of noise. PFT  \citep{zhao2022short} introduces a learnable enhancement patch to compensate for the problem of local feature extraction  \citep{islam2022recent} by transformer. Through feature slicing, merging, and stitching, the patch sequence can learn the long-range correlation of regions while ensuring the receptive field and paying more attention to local features. 
FED  \citep{wang2022feature} divides occlusion into non-pedestrian and non-target pedestrian occlusion, and focuses on the latter. Specifically, an occlusion set is first created to enhance the data. Then, a multiplicative occlusion score is introduced to diffuse the visible pedestrian parts, which improves the quality of the synthesized non-target pedestrians. The joint optimization of feature erasure and feature diffusion modules realizes the perception ability of the model to the target pedestrian.

FRT  \citep{xu2022learning} first classifies the pedestrian into head, torso, and legs using a pretrained pose estimation model HRNet  \citep{sun2019deep}, and extracts the corresponding features. 
Then the occlusion elimination module based on graph matching is introduced to eliminate the interference caused by occlusion by learning the similarity of common regions. 
Finally, FRT \citep{xu2022learning} recovers query features by aggregating neighbor features to solve the problem of information loss caused by occlusion. 
TL-TransNet  \citep{wang2022swin} uses a well-improved Swin transformer  \citep{liu2021swin} as a benchmark model to capture the main part of the person and uses DeepLabV3+  \citep{chen2018encoder} to remove background interference in the query and gallery. Finally, a reordering method based on hybrid similarity and background adaptation is designed to achieve the fusion of original features and removed background features. 
PADE  \citep{huang2022parallel} first obtains two new images through cropping and erasing operations and feeds them together with the original image into a multi-branch parameter sharing network with a vit  \citep{dosovitskiy2020image} as the backbone to enhance the global and local features in a manner similar to the self-attention mechanism.  
Finally, they are connected to form the final feature representation.

\subsection{ \textbf{Composite Method} }
By introducing more than two different networks, the occlusion problem is dealt with in an interactive or fused manner.

FGMFN  \citep{zhang2022fine} introduces a dual branch network. The local feature branches are first passed through an affine transformation based on ResNet18. Then, input into ResNet50  \citep{he2016deep} to get the upper body features and divide it into three regions, while introducing the attention module based on the gate mechanism to remove noise interference. The global feature branch adopts a feature extraction scheme based on block partition, and finally, the features of the two branches are fused proportionally as the final feature.  
Pirt  \citep{ma2021pose} first pre-trains the HRNet   \citep{sun2019deep} pose estimation model on the COCO  \citep{lin2014microsoft} dataset, from which a two-branch structure is elicited. Intra-part features are guided by a carefully modified ResNet50  \citep{he2016deep}, inter-part relationships are guided by a transformer, and the effects of occlusion are removed by establishing part-aware long-range dependencies.
Pirt  \citep{ma2021pose}  learns inter- and intra-part relationships in a collaborative manner.
MAT  \citep{zhou2022motion} first uses a CNN  to extract features from the image. Then, it flattens and passes them to a transformer. This method uses the methods of feature map and key point embedding to obtain a key point heat map and segmentation map. It uses affine transformation to obtain motion information of each key point and combines action information to realize fine-grained human body segmentation, and extracts representative pedestrian body parts to remove interference noise.  
DRL-Net  \citep{jia2022learning} first acquires obstacles from the training images to construct augmented samples with random obstacles. Then, a CNN and Transformer are concatenated to design a query-based semantic feature extraction layer. Finally, the semantic bootstrap is used to learn positive and negative sample comparisons and remove interference noise.

\subsection{ \textbf{3D Re-ID} }
Compared to 2D, the use of 3D information for occlusion person Re-ID is a relatively new approach. The shape and spatial depth features represented by 3D data are robust to texture information, and it usually reduces the interference caused by occlusion through 3D feature denoising, 3D feature complementation and multi-view construction.

PersonX  \citep{sun2019dissecting} constructs virtual 3D models by scanning people and objects in the real world and then maps them back to 2D for re-representation. Such manipulation of the data enhances the data representation.
 \citep{wang2022cloning} used UV texture mapping to clone clothing from real-world pedestrians to virtual 3D characters, while using a patch-based feature segmentation and expansion approach to deal with occlusion.
A more common form of 3D information is the point cloud  \citep{qi2017pointnet}, where the depth information in the point cloud can be used as an additional channel to the image.
OG-Net  \citep{zheng2022parameter} uses Skinned Multi-Person Linear ( SMPL   \citep{kanazawa2018end} ) to generate six channels of point cloud data from 2D images, providing positional and texture information.

ASSP  \citep{chen2021learning} uses 3D body reconstruction as an auxiliary task for 2D feature extraction. 
Specifically, texture-insensitive 3D shape information is first extracted from 2D images as auxiliary features. 
After that, a 3D row human model is created using SMPL  \citep{loper2015smpl}, while an adversarial learning and self-supervised projection combination is designed to combine 2D and 3D information into a 3D model for reconstruction \citep{ning2020multi}.
Regularization based on 3D reconstruction forces the model to decouple 3D shape information from visual features and remove the interference of noise to extract more discriminative features.
JGCL  \citep{chen2021joint} mitigates the lack of information caused by perspective in an unsupervised mapping.
 Specifically, a corresponding 3D mesh is first generated using a 3D network generator HMR  \citep{kanazawa2018end}.  
 After that, the model is rotated and combined with a GAN to generate diverse views of the human body.  
 These generated new views are combined with the original images in contrast learning while enhancing the view-invariant representation to improve the generated picture quality. 
3DT  \citep{zhang2022modeling} implements group person Re-ID using a 3D transformer. Because pedestrians in a group inevitably occlude each other, it transforms the 3D space into a discrete space by introducing a spatial layout token based on quantization and sampling, and it later assigns layout features to each member to reconstruct the spatial relationships between members.

However, the research on recognition using point clouds is still limited compared to 2D images, which is an important research direction for the future.

\subsection{ \textbf{Multi-modal Re-ID} }
\textbf{RGB-IR multimodal Re-ID.}  Both day and night are important scenes of pedestrian life, and in the case of insufficient illumination, images can only be collected by infrared cameras  \citep{nguyen2017person, wu2017rgb}. 
If there is occlusion in the scene, the infrared image still has occlusion.  
At the same time, the infrared image can be used as a special channel of the RGB image, which makes the representation mode of the RGB image more complete and can supplement the information well in the case of information loss caused by occlusion. 
 The RGB-IR multimodalRe-ID is designed to feed both infrared and conventional images into the model. 
 Removal of interference from occlusion is achieved by the relationship of different modalities or by combining approaches, such as attention mechanisms, in dealing with single modal noise, multi-scale, etc.

DDAG  \citep{ye2020dynamic} proposes a dynamic dual attention cross-modal graph structure.  
First, local attention is generated according to the similarity between features.  
Then, an aggregation representation of part-level relation learning is introduced.  
At the same time, the graph structure is introduced to remove the interference of noise based on relation. 
MSPAC  \citep{zhang2021multi} proposes a multi-scale-based component awareness mechanism. By adding an attention mechanism at the single scale, more fine-grained features are extracted in channel and spatial dimensions. After that, feature aggregation is realized using a cascade framework. By adding different scale features, the global structured body information is represented uniformly.
CM-LSP-GE  \citep{wang2022cross} is a framework for global features and local characteristics.  
The occlusion problem is solved by local features, and the alignment problem is solved by image segmentation and computing the shortest path of local features in two images.

To solve the intra-modal problem, HMML  \citep{zhang2022hybrid} introduces a pairing-based intra-modal similarity constraint to enhance the features. 
Similarly, CMC  \citep{wen2022cross} introduces multi-scale, multi-level feature learning to achieve refined feature extraction. To solve the problem of pedestrian pose misalignment caused by occlusion in transmembrane state, DAPN  \citep{liu2022dual} proposes an alignment network that can learn global and local modal invariance. Specifically,  DAPN   \citep{liu2022dual} first introduces an adaptive spatial transform module to align images. Second, the image is segmented horizontally to extract local features. Meanwhile, to learn the similarity representation of different modes, the different modes are embedded into a unified feature space \citep{cai2021dual}. Finally, the global and local features are fused as the final features. To solve the alignment problem, SPOT  \citep{chen2022structure} proposes a transformer-based network. First, a relational network composed of four convolution layers and two pooling layers is used to process the human body key point heat map to obtain the pedestrian body structure information. 
Combining it with the attention mechanism, the pedestrian region is highlighted to remove the background interference. 
Then, the transformer is used to mine the relationship between parts’ positions and features to enhance the discriminability of local features. The final feature is extracted by adding weight combination.
VI  \citep{park2021learning} first extracts features of IR and RGB using a dual-stream CNN. The cosine similarity of all corresponding feature pairs is calculated simultaneously, and the corresponding matching probability is calculated by the SoftMax function. These probabilities are taken as semantic similarity features of different modalities. Then, more discriminative features are extracted by local feature association. The above features are then aggregated to form the final feature. To solve the problem of image internal misalignment, DTRM  \citep{ye2021dynamic} combines attention and partial aggregation, and uses the context relationship of the two modalities to improve the global features to eliminate the noise effect.

\textbf{RGB-Depth multimodal Re-ID. }Depth  \citep{wang2021brief} images are captured through devices such as laser radars, and they provide body shape and skeletal information by measuring distance. For images, when information is lost owing to obstacles, depth features can supplement the position information of texture features to assist in a more complete expression. At the same time, it can solve the lighting problem and the problem of changing clothes for pedestrians. It can also help solve the problems of obstacle  illumination and different changing environments for clothing. RGB-Depth multimodal Re-ID aims to input depth map and RGB image simultaneously, and remove the interference of noise by the attention mechanism and other methods.

CMD  \citep{hafner2022cross} proposes an approach combining embedding representation and feature distillation. It removes noise through a gate-controlled attention mechanism. This mechanism uses one modality to dynamically activate the more discriminative features in another modality by gating signals.

\textbf{RGB-Text multimodal Re-ID.} 
RGB-Text multimodal Re-ID aims to introduce text data to enhance feature representation by sharing semantic information and attentional calibration to eliminate the effect of noise.In daily life, text information is one of the most frequently used types of information. When image information is missing or cannot be used owing to obstacles, it can be supplemented with text.

AXM-Net  \citep{farooq2022axm} dynamically exploits multi-scale information of text and images, recalibrating each modality according to the shared semantics and adding contextual attention to the text branch to supplement the information of the convolution block. 
Furthermore, attention is introduced to enhance feature consistency and local information of the visual part.  
It can learn the alignment semantic information of different modalities and automatically remove the interference of irrelevant information.

\section{ Method Comparison }
\begin{table*}[!t]
\caption{\label{table2} Comparison of experimental results based on local feature learning methods. 
The red numbers indicate the best results. (in $\%$).}
\centering
\resizebox{174mm}{40mm}{
\begin{tabular}{c|c|cccccccccccccc} 
\hline
\multicolumn{1}{c}{}                                                                 & \multicolumn{1}{c}{}                                                             & \multirow{2}{*}{Method} & \multirow{2}{*}{Venue} & \multicolumn{2}{c}{Occluded-Duke} & \multicolumn{2}{c}{Occluded-REID} & \multicolumn{2}{c}{Partial-REID} & \multicolumn{2}{c}{Partial-iLIDS} & \multicolumn{2}{c}{Market1501} & \multicolumn{2}{c}{DukeMTMC-reID}  \\
\multicolumn{1}{c}{}                                                                 & \multicolumn{1}{c}{}                                                             &                         &                        & Rank-1 & mAP                      & Rank-1 & mAP                      & Rank-1 & Rank-3                  & Rank-1 & Rank-3                   & Rank-1 & mAP                   & Rank-1 & mAP                       \\ 
\hline
\multirow{26}{*}{\begin{tabular}[c]{@{}c@{}}Local \\Feature \\Learning\end{tabular}} & \multirow{2}{*}{\begin{tabular}[c]{@{}c@{}}Body \\Segmentation\end{tabular}}     & CBDB-Net \citep{tan2021incomplete}                & TCSVT2021              & 50.09  & 38.9                     & -      & -                        & 66.7   & 78.3                    & 68.4   & 81.5                     & 94.4   & 85                    & 87.7   & 74.3                      \\
                                                                                     &                                                                                  & OCNet \citep{kim2022occluded}                  & ICASSP2022             & \textcolor{red}{\textbf{64.30}}  & \textcolor{red}{\textbf{54.40}}                   & -      & -                        & -      & -                       & -      & -                        & 95     & 89.3                  & 90.5   & 80.2                      \\ 
\cline{2-16}
                                                                                     & \multirow{11}{*}{\begin{tabular}[c]{@{}c@{}}Pose \\Estimation\end{tabular}}      & AACN \citep{xu2018attention}                    & CVPR2018               & -      & -                        & -      & -                        & -      & -                       & -      & -                        & 88.69  & 82.96                 & 76.84  & 59.25                         \\
                                                                                     &                                                                                  & ACSAP \citep{he2021adversarial}                   & ICIP2021               & -      & -                        & -      & -                        & 77     & 83.7                    & 76.5   & 87.4                     & -      & -                     & -      & -                         \\
                                                                                     &                                                                                  & DAReID \citep{xu2021dual}                  & KBS2021                & 63.4   & -                        & -      & -                        & 68.1   & 79.5                    & 76.7   & 85.3                     & 94.6   & 87                    & 88.9   & 78.4                      \\
                                                                                     &                                                                                  & DSA-reID \citep{zhang2019densely}                & CVPR2019               & -      & -                        & -      & -                        & -      & -                       & -      & -                        & \textcolor{red}{\textbf{95.7}}   & 87.6                  & 86.2   & 74.3                      \\
                                                                                     &                                                                                  & HOReID \citep{wang2020high}                  & CVPR2020               & -      & -                        & 80.3   & 70.2                     & 85.3   & 91                      & 72.6   & 86.4                     & 94.2   & 84.9                  & 86.9   & 75.6                      \\
                                                                                     &                                                                                  & PAFM \citep{yang2022pafm}                    & NCA2022                & 55.1   & 42.3                     & 76.4   & 68                       & 82.5   & -                       & -      & -                        & 95.6   & 88.5                  & \textcolor{red}{\textbf{91.2}}   & 80.1                      \\
                                                                                     &                                                                                  & PDC \citep{su2017pose}                     & ICCV2017               & -      & -                        & -      & -                        & -      & -                       & -      & -                        & 84.14  & 63.41                 & -      & -                         \\
                                                                                     &                                                                                  & PGFL-KD \citep{zheng2021pose}                 & MM2021                 & 63     & 54.1                     & 80.7   & 70.3                     & 85.1   & 90.8                    & 74     & 86.7                     & 95.3   & 87.2                  & 89.6   & 79.5                      \\
                                                                                     &                                                                                  & PGMANet \citep{zhai2021pgmanet}                 & IJCNN2021              & -      & -                        & -      & -                        & 82.1   & 85.5                    & 68.8   & 78.1                     & -      & -                     & -      & -                         \\
                                                                                     &                                                                                  & PMFB \citep{miao2021identifying}                    & TNNLS2021              & 56.3   & 43.5                     & -      & -                        & 72.5   & 83                      & 70.6   & 81.3                     & 92.7   & 81.3                  & 86.2   & 72.6                      \\
                                                                                     &                                                                                  & PVPM \citep{gao2020pose}                    & CVPR2020               & -      & -                        & 70.4   & 61.2                     & 78.3   & -                       & -      & -                        & -      & -                     & -      & -                         \\ 
\cline{2-16}
                                                                                     & \multirow{6}{*}{\begin{tabular}[c]{@{}c@{}}Semantic \\Segmentation\end{tabular}} & Co-Attention \citep{lin2021self}            & ICIP2021               & -      & -                        & -      & -                        & 83     & 90.3                    & 73.1   & 83.2                     & -      & -                     & -      & -                         \\
                                                                                     &                                                                                  & HPNet \citep{huang2020human}                   & ICME2020               & -      & -                        & \textcolor{red}{\textbf{87.3}}   & \textcolor{red}{\textbf{77.4}}                     & 85.7   & -                       & 68.9   & 80.7                     & -      & -                     & -      & -                         \\
                                                                                     &                                                                                  & MMGA \citep{cai2019multi}                    & CVPR2019               & -      & -                        & -      & -                        & -      & -                       & -      & -                        & 95     & 87.2                  & 89.5   & 78.1                      \\
                                                                                     &                                                                                  & SGSFA \citep{ren2020semantic}                   & PMLR2020               & 62.3   & 47.4                     & 63.1   & 53.2                     & 68.2   & -                       & -      & -                        & 92.3   & 80.2                  & 84.7   & 70.8                      \\
                                                                                     &                                                                                  & SORN \citep{zhang2020semantic}                    & TCSVT2020              & 57.6   & 46.3                     & -      & -                        & 76.7   & 84.3                    & 79.8   & 86.6                     & 94.8   & 84.5                  & 86.9   & 74.1                      \\
                                                                                     &                                                                                  & SPReID \citep{kalayeh2018human}                  & CVPR2017               & -      & -                        & -      & -                        & -      & -                       & -      & -                        & 94.63  & \textcolor{red}{\textbf{90.96}}                 & 88.96  & \textcolor{red}{\textbf{84.99}}                     \\ 
\cline{2-16}
                                                                                     & \begin{tabular}[c]{@{}c@{}}Attribute \\Annotation\end{tabular}                   & ASAN \citep{jin2021occlusion}                    & TCSVT2021              & 55.40  & 43.80                    & 82.50  & 71.80                    & \textcolor{red}{\textbf{86.80}}  & 93.50                   & \textcolor{red}{\textbf{81.70}}  & \textcolor{red}{\textbf{88.30}}                    & 94.60  & 85.30                 & 87.50  & 76.30                     \\ 
\cline{2-16}
                                                                                     & \multirow{6}{*}{\begin{tabular}[c]{@{}c@{}}Mixing \\Method\end{tabular}}         & FGSA \citep{zhou2020fine}                    & TIP2020                & -      & -                        & -      & -                        & -      & -                       & -      & -                        & 91.50  & 85.40                 & 85.90  & 74.10                     \\
                                                                                     &                                                                                  & GASM \citep{he2020guided}                    & ECCV2020               & -      & -                        & 80.30  & 73.10                    & -      & -                       & -      & -                        & 95.30  & 84.70                 & 88.30  & 74.40                     \\
                                                                                     &                                                                                  & PGFA \citep{miao2019pose}                    & ICCV2019               & -      & -                        & -      & -                        & 68.80  & 80.00                   & 69.10  & 80.90                    & 91.20  & 76.80                 & 82.60  & 65.50                     \\
                                                                                     &                                                                                  & SSPReID \citep{quispe2019improved}                & IVC 2019               & -      & -                        & -      & -                        & -      & -                       & -      & -                        & 93.70  & 90.80                 & 86.40  & 83.70                     \\
                                                                                     &                                                                                  & LKWS \citep{yang2021learning}                    & ICCV2021               & 62.2   & 46.3                     & 81     & 71                       & 85.7   & \textcolor{red}{\textbf{93.7}}                    & 80.7   & 88.2                     & -      & -                     & -      & -                         \\
                                                                                     &                                                                                  & TSA \citep{gao2020texture}                     & ACM MM2020             & -      & -                        & -      & -                        & 72.70  & 85.20                   & 73.90  & 84.70                    & -      & -                     & -      & -                         \\
\hline
\end{tabular}}
\end{table*}

\begin{table*}[!t]
\caption{\label{table3} Comparison of experimental results based on relationship representation methods. 
The red numbers indicate the best results.(in $\%$).}
\centering
\resizebox{174mm}{32mm}{
\begin{tabular}{c|c|cccccccccccccc} 
\hline
\multicolumn{1}{c}{}                                                                    & \multicolumn{1}{c}{}                                                   &              &            & \multicolumn{2}{c}{Occluded-Duke} & \multicolumn{2}{c}{Occluded-REID} & \multicolumn{2}{c}{Partial-REID} & \multicolumn{2}{c}{Partial-iLIDS} & \multicolumn{2}{c}{Market1501} & \multicolumn{2}{c}{DukeMTMC-reID}  \\
\multicolumn{1}{c}{}                                                                    & \multicolumn{1}{c}{}                                                   & Method       & Venue      & Rank-1 & mAP                      & Rank-1 & mAP                      & Rank-1 & Rank-3                  & Rank-1 & Rank-3                   & Rank-1 & mAP                   & Rank-1 & mAP                       \\ 
\hline
\multirow{18}{*}{\begin{tabular}[c]{@{}c@{}}Relationship \\Representation\end{tabular}} & \multirow{3}{*}{\begin{tabular}[c]{@{}c@{}}Shared \\Area\end{tabular}} & KBFM \citep{han2020keypoint}         & ICIP2020   & -      & -                        & -      & -                        & 69.7   & 82.2                    & 64.1   & 73.9                     & -      & -                     & -      & -                         \\
                                                                                        &                                                                        & PPCL \citep{he2021partial}         & CVPR2021   & -      & -                        & -      & -                        & 83.70  & 88.70                   & 71.40  & 85.70                    & -      & -                     & -      & -                         \\
                                                                                        &                                                                        & VPM \citep{sun2019perceive}          & CVPR2019   & -      & -                        & -      & -                        & 67.70  & 81.90                   & 65.50  & 74.80                    & 90.40  & 75.70                 & -      & -                         \\ 
\cline{2-16}
                                                                                        & Clustering                                                             & ISP \citep{zhu2020identity}          & ECCV2020   & 62.80  & 52.30                    & -      & -                        & -      & -                       & -      & -                        & 94.63  & 90.69                 & 88.96  & 84.99                     \\ 
\cline{2-16}
                                                                                        & \begin{tabular}[c]{@{}c@{}}Figure \\Convolution\end{tabular}           & HOReID \citep{wang2020high}       & CVPR2020   & -      & -                        & \textcolor{red}{\textbf{80.3}}   & \textcolor{red}{\textbf{70.2}}                     & 85.3   & 91                      & 72.6   & 86.4                     & 94.2   & 84.9                  & 86.9   & 75.6                      \\ 
\cline{2-16}
                                                                                        & \multirow{13}{*}{Attention}                                            & AACN \citep{xu2018attention}         & CVPR2018   & -      & -                        & -      & -                        & -      & -                       & -      & -                        & 88.69  & 82.96                 & 76.84  & 59.25                         \\
                                                                                        &                                                                        & APN \citep{huo2021attentive}          & ICPR2021   & -      & -                        & -      & -                        & 71.80  & 85.50                   & 66.40  & 76.50                    & \textcolor{red}{\textbf{96.00}}  & 89.00                 & 89.50  & 79.20                     \\
                                                                                        &                                                                        & CASN \citep{zheng2019re}         & CVPR2019   & -      & -                        & -      & -                        & -      & -                       & -      & -                        & 94.40  & 82.80                 & 87.70  & 73.70                     \\
                                                                                        &                                                                        & Co-Attention \citep{lin2021self} & ICIP2021   & -      & -                        & -      & -                        & 83.00  & 90.30                   & 73.10  & 83.20                    & -      & -                     & -      & -                         \\
                                                                                        &                                                                        & DAReID \citep{xu2021dual}       & KBS2021    & 63.4   & -                        & -      & -                        & 68.1   & 79.5                    & 76.7   & 85.3                     & 94.6   & 87                    & 88.9   & 78.4                      \\
                                                                                        &                                                                        & DSOP \citep{wang2020deep}         & JPCS2020   & 57.70  & 45.30                    & -      & -                        & -      & -                       & -      & -                        & 95.40  & 85.90                 & 88.20  & 77.00                     \\
                                                                                        &                                                                        & MHSA-Net \citep{tan2022mhsa}     & TNNLS2022  & 59.70  & 44.80                    & -      & -                        & \textcolor{red}{\textbf{85.70}}  & \textcolor{red}{\textbf{91.30}}                   & 74.90  & \textcolor{red}{\textbf{87.20}}                    & 95.50  & \textcolor{red}{\textbf{93.00}}                 & 90.70  & \textcolor{red}{\textbf{87.20}}                     \\
                                                                                        &                                                                        &OCNet \citep{kim2022occluded}       & ICASSP2022 & 64.30& \textcolor{red}{\textbf{54.40}}                    & -      & -                        & -      & -                       & -      & -                        & -      & -                     & -      & -                         \\
                                                                                        &                                                                        & PAFM \citep{yang2022pafm}         & NCA2022    & 55.1   & 42.3                     & 76.4   & 68                       & 82.5   & -                       & -      & -                        & 95.6   & 88.5                  & \textcolor{red}{\textbf{91.2}}   & 80.1                      \\
                                                                                        &                                                                        & PISNet \citep{zhao2020not}       & ECCV2020   & -      & -                        & -      & -                        & -      & -                       & -      & -                        & 95.60  & 87.10                 & 88.80  & 78.70                     \\
                                                                                        &                                                                        & PSE \citep{sarfraz2018pose}          & CVPR2018   & -      & -                        & -      & -                        & -      & -                       & -      & -                        & 90.30  & 84.00                 & 85.20  & 79.80                     \\
                                                                                        &                                                                        & QPM \citep{wang2022quality}          & ITM2022    & \textcolor{red}{\textbf{64.40}}  & 49.70                    & -      & -                        & 81.70  & 88.00                   & \textcolor{red}{\textbf{77.30}}  & 85.70                    & -      & -                     & -      & -                         \\
                                                                                        &                                                                        & VPM \citep{sun2019perceive}          & CVPR2019   & -      & -                        & -      & -                        & 67.70  & 81.90                   & 65.50  & 74.80                    & 90.40  & 75.70                 & -      & -                         \\
\hline
\end{tabular}}
\end{table*}

\begin{table*}[!t]
\caption{\label{table4} Comparison of experimental results with other methods. 
The red numbers indicate the best results.(in $\%$).}
\centering
\resizebox{174mm}{30mm}{
\begin{tabular}{c|c|cccccccccccccc} 
\hline
\multicolumn{1}{c}{}                                                                &            &             & \multicolumn{2}{c}{Occluded-Duke} & \multicolumn{2}{c}{Occluded-REID} & \multicolumn{2}{c}{Partial-REID} & \multicolumn{2}{c}{Partial-iLIDS} & \multicolumn{2}{c}{Market1501} & \multicolumn{2}{c}{DukeMTMC-reID}  \\
\multicolumn{1}{c}{}                                                                & Method     & Venue       & Rank-1 & mAP                      & Rank-1 & mAP                      & Rank-1 & Rank-3                  & Rank-1 & Rank-3                   & Rank-1 & mAP                   & Rank-1 & mAP                       \\ 
\hline
\multirow{6}{*}{Transformer}                                                        & DPM \citep{tan2022dynamic}        & ACM MM 2022 & \textcolor{red}{\textbf{71.40}}  & \textcolor{red}{\textbf{61.80}}                    & 85.50  & 79.70                    & -      & -                       & -      & -                        & 95.50  & 89.70                 & 91.00  & 82.60                     \\
                                                                                    & FED \citep{wang2022feature}        & CVPR2022    & 68.10  & 56.40                    & \textcolor{red}{\textbf{86.30}}  & 79.30                    & 83.10  & -                       & -      & -                        & 95.00  & 86.30                 & 89.40  & 78.00                     \\
                                                                                    & FRT \citep{xu2022learning}        & TIP2022     & 70.70  & 61.30                    & 80.40  & 71.00                    & \textcolor{red}{\textbf{88.20}}  & \textcolor{red}{\textbf{93.20}}                  & 73.00  & 87.00                    & 95.50  & 88.10                 & 90.50  & 81.70                     \\
                                                                                    & TransReID \citep{he2021transreid} & ICCV2021    & 66.40  & 59.20                    & -      & -                        & -      & -                       & -      & -                        & 95.20  & 88.90                 & 90.70  & 82.00                     \\
                                                                                    & PFD \citep{wang2022pose}        & AAAI2022    & 69.50  & \textcolor{red}{\textbf{61.80}}                    & 81.50  & \textcolor{red}{\textbf{83.00}}                    & -      & -                       & -      & -                        & 95.50  & 89.70                 & \textcolor{red}{\textbf{91.20}}  & 83.20                     \\
                                                                                    & PFT \citep{zhao2022short}        & NCA2022     & 69.80  & 60.80                    & 83.00  & 78.30                    & 81.30  & 79.90                   & 68.10  & 81.50                    & 95.30  & 88.80                 & 90.70  & 82.10                     \\ 
\hline
\multirow{2}{*}{\begin{tabular}[c]{@{}c@{}}Mixing \\Method\end{tabular}}            & DRL-Net \citep{jia2022learning}    & TMM2021     & 65.80  & 53.90                    & -      & -                        & -      & -                       & -      & -                        & 94.70  & 86.90                 & 88.10  & 76.60                     \\
                                                                                    & Pirt \citep{ma2021pose}       & ACMMM2021   & 60.00  & 50.90                    & -      & -                        & -      & -                       & -      & -                        & 94.10  & 86.30                 & 88.90  & 77.60                     \\ 
\hline
\multirow{2}{*}{Multiscale}                                                         & DSR \citep{he2018deep}        & CVPR2018    & 40.80  & 30.40                    & 72.80  & 62.80                    & 50.70  & 70.00                   & 58.80  & 67.20                    & 83.58  & 64.25                 & -      & -                         \\
                                                                                    & FPR \citep{he2019foreground}        & ICCV2019    & -      & -                        & 78.30  & 68.00                    & 81.00  & -                       & 68.10  & -                        & 95.42  & 86.58                 & 88.64  & 78.42                     \\ 
\hline
\multirow{2}{*}{\begin{tabular}[c]{@{}c@{}}Regional \\Reconfiguration\end{tabular}} & ACSAP \citep{he2021adversarial}      & ICIP2021    & -      & -                        & -      & -                        & 77.00  & 83.70                   & 76.50  & \textcolor{red}{\textbf{87.40}}                    & -      & -                     & -      & -                         \\
                                                                                    & RFCNet \citep{hou2021feature}    & TPAMI2021   & 63.90  & 54.50                    & -      & -                        & -      & -                       & -      & -                        & 95.20  & 89.20                 & 90.70  & 80.70                     \\ 
\hline
\multirow{3}{*}{\begin{tabular}[c]{@{}c@{}}Data \\Enhancement\end{tabular}}         & IGOAS \citep{zhao2021incremental}      & TIP2021     & 60.10  & 49.40                    & 81.10  & -                        & -      & -                       & -      & -                        & 93.40  & 84.10                 & 86.90  & 75.10                     \\
                                                                                    & OAMN \citep{chen2021occlude}       & ICCV2021    & 62.60  & 46.10                    & -      & -                        & 86.00  & -                       & \textcolor{red}{\textbf{77.30}}  & -                        & 93.20  & 79.80                 & 86.30  & 72.60                     \\
                                                                                    & SSGR \citep{yan2021occluded}       & ICCV2021    & 65.80  & 57.20                    & 78.50  & 72.90                    & -      & -                       & -      & -                        & \textcolor{red}{\textbf{96.10}}  & 89.30                 & 91.10  & 81.30                     \\ 
\hline
Regularization                                                                      & MHSA-Net \citep{tan2022mhsa}   & TNNLS2022   & 59.70  & 44.80                    & -      & -                        & 85.70  & 91.30                   & 74.90  & 87.20                    & 95.50  & \textcolor{red}{\textbf{93.00}}                 & 90.70  & \textcolor{red}{\textbf{87.20}}                     \\
\hline
\end{tabular}}
\end{table*}

We statistically evaluate the results of the occluded person Re-ID methods on two general datasets (Market1501  \citep{zheng2015scalable}, DukeMTMC-reID  \citep{ristani2016performance}), two occluded person Re-ID datasets (Occluded-DukeMTMC  \citep{miao2019pose}, Occluded-REID  \citep{zheng2015partial}) and two partial person Re-ID datasets (Partial-ReID  \citep{zheng2015partial}, Partial-iLIDS  \citep{he2018deep}). We simply divide them into three categories: 
The experimental results based on the local feature learning method are shown in Table \ref{table2}. It contains body segmentation, pose estimation, semantic segmentation, attribute annotation, and mixing method. The experimental results based on the relationship representation method are shown in Table \ref{table3}. 
It contains shared area, clustering, figure convolution, and attention. The experimental results of other methods are shown in Table \ref{table4}. It contains the transformer, mixing method, multi-scale, regional reconfiguration, data enhancement, and regularization. For an introduction to each class of methods and details of each study, the reader is referred to Section 4.

From the results we can get the following information:
(1) From the results we can get the following information:
OCNet  \citep{kim2022occluded} based on local feature learning, QPM  \citep{wang2022quality} based on relational representation, DPM  \citep{tan2022dynamic} and PFD  \citep{wang2022pose} based on transformer perform better on the Occluded-DukeMTMC dataset.
On the Occluded-REID dataset, HPNet  \citep{huang2020human} based on local feature learning, HOReID  \citep{wang2020high} based on relational representation, FED  \citep{wang2022feature} and PFD  \citep{wang2022pose} based on transformer perform stably.
On the Partial-ReID dataset, ASAN  \citep{jin2021occlusion} and LKWS  \citep{yang2021learning} based on local feature learning, MHSA-Net  \citep{tan2022mhsa} based on relational representation, and FRT  \citep{xu2022learning} based on transformer achieve better performance.
On the Partial-iLIDS dataset,  ASAN  \citep{jin2021occlusion} based on local feature learning, MHSA-Net  \citep{tan2022mhsa} and QPM  \citep{wang2022quality} based on relational representation, ACSAP  \citep{he2021adversarial} based on region reconstruction, and OAMN  \citep{chen2021occlude} based on data augmentation achieve very stable performance. 
Therefore, there is no single method that achieves the best performance on all datasets.
(2) In general, the performance on the occlusion dataset reflects the ability of the model to resist noise, the performance on the partial dataset reflects the recognition ability of the model under the condition of missing pedestrian information, and the performance on the general dataset reflects the comprehensive performance of the model. 
Each of these approaches addresses one or more specific problems. 
(3) Attention and pose estimation are the more mainstream and typical of the many pedestrian re-identification methods for dealing with occlusion.
Attribute annotation-based, clustering-based, figure convolution-based and regularisation-based methods, on the other hand, have received less attention.

\section{ Future directions }
\subsection{ \textbf{Richer, higher quality datasets} }
Most models are evaluated on datasets collected in controlled environments.
The data in real-world scenarios is often uncontrollable, which can significantly impact the model's performance on such datasets.

For an occluded person Re-ID dataset, it is necessary to incorporate one or more modal inputs, including a diverse combination of images, text, infrared maps, and depth maps. This enables the model to effectively handle a wider range of realistic occlusion scenarios.
Moreover, there is a lack of larger datasets encompassing a broader range of domains, environments  \citep{gou2018systematic}, and higher resolutions, which would provide richer content and higher quality for research purposes.

\subsection{ \textbf{More robust and varied feature extraction} }
\textbf{3D Re-ID.} 
Inspired by human three-dimensional cognition, some researchers believe that the complete pedestrian representation should be a fusion of 3D and 2D \citep{zheng2022parameter}.

At present, PointNet  \citep{qi2017pointnet}, as a representative of deep learning methods in point cloud feature extraction  \citep{wang2022learning, wang20223d}, has demonstrated promising results.
Point cloud completion  \citep{fei2022comprehensive} and point cloud correction can aid in 3D occluded person Re-ID, while 3D pose estimation  \citep{wang2021deep} and 3D semantic segmentation  \citep{xie2020linking} can guide the feature extraction process for person Re-ID. However, research in the 3D domain for pedestrian recognition  \citep{zhao2013unsupervised, sun2019perceive} is still relatively limited compared to the advancements in 2D approaches. Therefore, 3D occluded person Re-ID represents a significant and promising research direction for the future  \citep{tirkolaee2020fuzzy}.

\textbf{Multimodal Re-ID.}  
The information captured from different modalities demonstrates a significant diversity in content representation \citep{wu2017rgb, wu2017robust}. Improving the interaction, fusion, and extraction of more comprehensive pedestrian features at both the data and feature extraction stages represents a crucial research direction for future advancements  \citep{tutsoy2022priority, sekhar2017conductive}.

\textbf{ Cross-resolution occluded person Re-ID. }
Owing to the influence of the distance and pixel size of the collection device, the resolution of the collected samples is uneven, and the feature space correspondence is also inconsistent  \citep{li2019recover}. 
At the same time, low resolution will lose significant spatial and detail information  \citep{mao2019resolution}. 
How to extract pedestrian features at different resolutions under occlusion conditions is a problem to be solved in the future.

{\textbf{ Fast response, smaller occlusion person Re-ID model.} 
Constructing smaller  \citep{zhou2019omni, li2018harmonious} and faster  \citep{liu2017neural} occluded person Re-ID models with reduced parameters is a crucial research direction for future advancements  \citep{tutsoy2018design}.

\textbf{ Unsupervised and semi-supervised occluded person Re-ID.  } 
The complex manual labeling process is omitted, and the pedestrian features are learned by using the datasets without labels  \citep{zhang2018deep, liu2019deep} or with a small number of labels  \citep{wang2019learning, wang2019rgb, nagaraju2016hierarchical}.

Currently, the performance of unsupervised and semi-supervised based methods on the task of occluded person Re-ID is still some way off compared to supervised methods. Supervised methods usually rely on large-scale labelled datasets for training and thus can achieve high performance. However, as unsupervised and semi-supervised methods are increasingly studied, they show significant potential in improving the generalisation of occluded person Re-ID models.

\subsection{ \textbf{Occluded person Re-ID system} }
At present, few researchers combine object detection and occluded person Re-ID together. The end-to-end person Re-ID systems are lacking, and the integrated system has more applications in real life  \citep{martinel2016temporal}. 
How to combine the two more effectively and rationally and design a occluded person Re-ID system that is more robust to occlusion is an important research direction.

\section{ Conclusion }
This review offers a comprehensive and integrated analysis and discussion of deep learning methods for occluded person Re-ID, addressing both practical and research-driven requirements. Firstly, we introduce the classification of occlusion problems and the datasets specifically designed for occluded person Re-ID. Secondly, we comprehensively classify and introduce the methods proposed in top international journals and conferences before 2023 for addressing occluded person Re-ID. Finally, the future prospects of occluded person Re-ID are analyzed from data, feature, and system perspectives, respectively. In this study, we categorize the most significant image feature extraction methods into five major categories: local feature learning, relational representation, transformer-based methods, mixing methods, and other approaches. This review will assist researchers in comprehending the process and objectives of these methods, providing valuable references and contributing to the research significance in the advancement of occluded Re-ID.


\bibliographystyle{apalike2}

\begin{thebibliography}{}

\bibitem[Bedagkar-Gala \& Shah, 2014]{bedagkar2014survey}
Bedagkar-Gala, A. \& Shah, S.~K. (2014).
\newblock A survey of approaches and trends in person re-identification.
\newblock {\em Image and vision computing}, 32(4), 270--286.

\bibitem[Cai et~al., 2019]{cai2019multi}
Cai, H., Wang, Z., \& Cheng, J. (2019).
\newblock Multi-scale body-part mask guided attention for person re-identification.
\newblock In {\em Proceedings of the IEEE/CVF Conference on Computer Vision and Pattern Recognition Workshops}  (pp.\ 0--0).

\bibitem[Cai et~al., 2021]{cai2021dual}
Cai, X., Liu, L., Zhu, L., \& Zhang, H. (2021).
\newblock Dual-modality hard mining triplet-center loss for visible infrared person re-identification.
\newblock {\em Knowledge-Based Systems}, 215, 106772.

\bibitem[Chen et~al., 2022]{chen2022structure}
Chen, C., Ye, M., Qi, M., Wu, J., Jiang, J., \& Lin, C.-W. (2022).
\newblock Structure-aware positional transformer for visible-infrared person re-identification.
\newblock {\em IEEE Transactions on Image Processing}, 31, 2352--2364.

\bibitem[Chen et~al., 2021a]{chen2021joint}
Chen, H., Wang, Y., Lagadec, B., Dantcheva, A., \& Bremond, F. (2021a).
\newblock Joint generative and contrastive learning for unsupervised person re-identification.
\newblock In {\em Proceedings of the IEEE/CVF conference on computer vision and pattern recognition}  (pp.\ 2004--2013).

\bibitem[Chen et~al., 2021b]{chen2021learning}
Chen, J., Jiang, X., Wang, F., Zhang, J., Zheng, F., Sun, X., \& Zheng, W.-S. (2021b).
\newblock Learning 3d shape feature for texture-insensitive person re-identification.
\newblock In {\em Proceedings of the IEEE/CVF Conference on Computer Vision and Pattern Recognition}  (pp.\ 8146--8155).

\bibitem[Chen et~al., 2018]{chen2018encoder}
Chen, L.-C., Zhu, Y., Papandreou, G., Schroff, F., \& Adam, H. (2018).
\newblock Encoder-decoder with atrous separable convolution for semantic image segmentation.
\newblock In {\em Proceedings of the European conference on computer vision (ECCV)}  (pp.\ 801--818).

\bibitem[Chen et~al., 2021c]{chen2021occlude}
Chen, P., Liu, W., Dai, P., Liu, J., Ye, Q., Xu, M., Chen, Q., \& Ji, R. (2021c).
\newblock Occlude them all: Occlusion-aware attention network for occluded person re-id.
\newblock In {\em Proceedings of the IEEE/CVF international conference on computer vision}  (pp.\ 11833--11842).

\bibitem[Cheng et~al., 2011]{cheng2011custom}
Cheng, D.~S., Cristani, M., Stoppa, M., Bazzani, L., \& Murino, V. (2011).
\newblock Custom pictorial structures for re-identification.
\newblock In {\em Bmvc}, volume~1  (pp.\~6).: Citeseer.

\bibitem[CS et~al., 2022]{wang20223d}
CS, W., H, W., X, N., SW, T., \& WJ, L. (2022).
\newblock 3d point cloud classification method based on dynamic coverage of local area.
\newblock {\em Journal of Software}, (pp.\ 0--0).

\bibitem[Dong et~al., 2023]{dong2023erasing}
Dong, N., Zhang, L., Yan, S., Tang, H., \& Tang, J. (2023).
\newblock Erasing, transforming, and noising defense network for occluded person re-identification.
\newblock {\em arXiv preprint arXiv:2307.07187}.

\bibitem[Dosovitskiy et~al., 2020]{dosovitskiy2020image}
Dosovitskiy, A., Beyer, L., Kolesnikov, A., Weissenborn, D., Zhai, X., Unterthiner, T., Dehghani, M., Minderer, M., Heigold, G., Gelly, S., et~al. (2020).
\newblock An image is worth 16x16 words: Transformers for image recognition at scale.
\newblock {\em arXiv preprint arXiv:2010.11929}.

\bibitem[Ess et~al., 2008]{ess2008mobile}
Ess, A., Leibe, B., Schindler, K., \& Van~Gool, L. (2008).
\newblock A mobile vision system for robust multi-person tracking.
\newblock In {\em 2008 IEEE Conference on Computer Vision and Pattern Recognition}  (pp.\ 1--8).: IEEE.

\bibitem[Farooq et~al., 2022]{farooq2022axm}
Farooq, A., Awais, M., Kittler, J., \& Khalid, S.~S. (2022).
\newblock Axm-net: Implicit cross-modal feature alignment for person re-identification.
\newblock In {\em Proceedings of the AAAI Conference on Artificial Intelligence}, volume~36  (pp.\ 4477--4485).

\bibitem[Fei et~al., 2022]{fei2022comprehensive}
Fei, B., Yang, W., Chen, W.-M., Li, Z., Li, Y., Ma, T., Hu, X., \& Ma, L. (2022).
\newblock Comprehensive review of deep learning-based 3d point cloud completion processing and analysis.
\newblock {\em IEEE Transactions on Intelligent Transportation Systems}.

\bibitem[Fu et~al., 2019]{fu2019dual}
Fu, J., Liu, J., Tian, H., Li, Y., Bao, Y., Fang, Z., \& Lu, H. (2019).
\newblock Dual attention network for scene segmentation.
\newblock In {\em Proceedings of the IEEE/CVF conference on computer vision and pattern recognition}  (pp.\ 3146--3154).

\bibitem[Gao et~al., 2020a]{gao2020texture}
Gao, L., Zhang, H., Gao, Z., Guan, W., Cheng, Z., \& Wang, M. (2020a).
\newblock Texture semantically aligned with visibility-aware for partial person re-identification.
\newblock In {\em Proceedings of the 28th ACM International Conference on Multimedia}  (pp.\ 3771--3779).

\bibitem[Gao et~al., 2020b]{gao2020pose}
Gao, S., Wang, J., Lu, H., \& Liu, Z. (2020b).
\newblock Pose-guided visible part matching for occluded person reid.
\newblock In {\em Proceedings of the IEEE/CVF conference on computer vision and pattern recognition}  (pp.\ 11744--11752).

\bibitem[Gou et~al., 2018]{gou2018systematic}
Gou, M., Wu, Z., Rates-Borras, A., Camps, O., Radke, R.~J., et~al. (2018).
\newblock A systematic evaluation and benchmark for person re-identification: Features, metrics, and datasets.
\newblock {\em IEEE transactions on pattern analysis and machine intelligence}, 41(3), 523--536.

\bibitem[G{\"u}ler et~al., 2018]{guler2018densepose}
G{\"u}ler, R.~A., Neverova, N., \& Kokkinos, I. (2018).
\newblock Densepose: Dense human pose estimation in the wild.
\newblock In {\em Proceedings of the IEEE conference on computer vision and pattern recognition}  (pp.\ 7297--7306).

\bibitem[Hafner et~al., 2022]{hafner2022cross}
Hafner, F.~M., Bhuyian, A., Kooij, J.~F., \& Granger, E. (2022).
\newblock Cross-modal distillation for rgb-depth person re-identification.
\newblock {\em Computer Vision and Image Understanding}, 216, 103352.

\bibitem[Han et~al., 2020]{han2020keypoint}
Han, C., Gao, C., \& Sang, N. (2020).
\newblock Keypoint-based feature matching for partial person re-identification.
\newblock In {\em 2020 IEEE International Conference on Image Processing (ICIP)}  (pp.\ 226--230).: IEEE.

\bibitem[He et~al., 2016]{he2016deep}
He, K., Zhang, X., Ren, S., \& Sun, J. (2016).
\newblock Deep residual learning for image recognition.
\newblock In {\em Proceedings of the IEEE conference on computer vision and pattern recognition}  (pp.\ 770--778).

\bibitem[He et~al., 2018]{he2018deep}
He, L., Liang, J., Li, H., \& Sun, Z. (2018).
\newblock Deep spatial feature reconstruction for partial person re-identification: Alignment-free approach.
\newblock In {\em Proceedings of the IEEE conference on computer vision and pattern recognition}  (pp.\ 7073--7082).

\bibitem[He \& Liu, 2020]{he2020guided}
He, L. \& Liu, W. (2020).
\newblock Guided saliency feature learning for person re-identification in crowded scenes.
\newblock In {\em Computer Vision--ECCV 2020: 16th European Conference, Glasgow, UK, August 23--28, 2020, Proceedings, Part XXVIII 16}  (pp.\ 357--373).: Springer.

\bibitem[He et~al., 2019]{he2019foreground}
He, L., Wang, Y., Liu, W., Zhao, H., Sun, Z., \& Feng, J. (2019).
\newblock Foreground-aware pyramid reconstruction for alignment-free occluded person re-identification.
\newblock In {\em Proceedings of the IEEE/CVF international conference on computer vision}  (pp.\ 8450--8459).

\bibitem[He et~al., 2021a]{he2021transreid}
He, S., Luo, H., Wang, P., Wang, F., Li, H., \& Jiang, W. (2021a).
\newblock Transreid: Transformer-based object re-identification.
\newblock In {\em Proceedings of the IEEE/CVF international conference on computer vision}  (pp.\ 15013--15022).

\bibitem[He et~al., 2021b]{he2021partial}
He, T., Shen, X., Huang, J., Chen, Z., \& Hua, X.-S. (2021b).
\newblock Partial person re-identification with part-part correspondence learning.
\newblock In {\em Proceedings of the IEEE/CVF Conference on Computer Vision and Pattern Recognition}  (pp.\ 9105--9115).

\bibitem[He et~al., 2021c]{he2021adversarial}
He, Y., Yang, H., \& Chen, L. (2021c).
\newblock Adversarial cross-scale alignment pursuit for seriously misaligned person re-identification.
\newblock In {\em 2021 IEEE International Conference on Image Processing (ICIP)}  (pp.\ 2373--2377).: IEEE.

\bibitem[Hou et~al., 2021]{hou2021feature}
Hou, R., Ma, B., Chang, H., Gu, X., Shan, S., \& Chen, X. (2021).
\newblock Feature completion for occluded person re-identification.
\newblock {\em IEEE Transactions on Pattern Analysis and Machine Intelligence}, 44(9), 4894--4912.

\bibitem[Huang et~al., 2020]{huang2020human}
Huang, H., Chen, X., \& Huang, K. (2020).
\newblock Human parsing based alignment with multi-task learning for occluded person re-identification.
\newblock In {\em 2020 IEEE international conference on multimedia and expo (ICME)}  (pp.\ 1--6).: IEEE.

\bibitem[Huang et~al., 2022]{huang2022parallel}
Huang, H., Zheng, A., Li, C., He, R., et~al. (2022).
\newblock Parallel augmentation and dual enhancement for occluded person re-identification.
\newblock {\em arXiv preprint arXiv:2210.05438}.

\bibitem[Huo et~al., 2021]{huo2021attentive}
Huo, L., Song, C., Liu, Z., \& Zhang, Z. (2021).
\newblock Attentive part-aware networks for partial person re-identification.
\newblock In {\em 2020 25th International Conference on Pattern Recognition (ICPR)}  (pp.\ 3652--3659).: IEEE.

\bibitem[Islam, 2022]{islam2022recent}
Islam, K. (2022).
\newblock Recent advances in vision transformer: A survey and outlook of recent work.
\newblock {\em arXiv preprint arXiv:2203.01536}.

\bibitem[Jia et~al., 2022]{jia2022learning}
Jia, M., Cheng, X., Lu, S., \& Zhang, J. (2022).
\newblock Learning disentangled representation implicitly via transformer for occluded person re-identification.
\newblock {\em IEEE Transactions on Multimedia}.

\bibitem[Jin et~al., 2021]{jin2021occlusion}
Jin, H., Lai, S., \& Qian, X. (2021).
\newblock Occlusion-sensitive person re-identification via attribute-based shift attention.
\newblock {\em IEEE Transactions on Circuits and Systems for Video Technology}, 32(4), 2170--2185.

\bibitem[Kalayeh et~al., 2018]{kalayeh2018human}
Kalayeh, M.~M., Basaran, E., G{\"o}kmen, M., Kamasak, M.~E., \& Shah, M. (2018).
\newblock Human semantic parsing for person re-identification.
\newblock In {\em Proceedings of the IEEE conference on computer vision and pattern recognition}  (pp.\ 1062--1071).

\bibitem[Kanazawa et~al., 2018]{kanazawa2018end}
Kanazawa, A., Black, M.~J., Jacobs, D.~W., \& Malik, J. (2018).
\newblock End-to-end recovery of human shape and pose.
\newblock In {\em Proceedings of the IEEE conference on computer vision and pattern recognition}  (pp.\ 7122--7131).

\bibitem[Khan et~al., 2022]{khan2022transformers}
Khan, S., Naseer, M., Hayat, M., Zamir, S.~W., Khan, F.~S., \& Shah, M. (2022).
\newblock Transformers in vision: A survey.
\newblock {\em ACM computing surveys (CSUR)}, 54(10s), 1--41.

\bibitem[Kim \& Yoo, 2017]{kim2017deep}
Kim, J. \& Yoo, C.~D. (2017).
\newblock Deep partial person re-identification via attention model.
\newblock In {\em 2017 IEEE International Conference on Image Processing (ICIP)}  (pp.\ 3425--3429).: IEEE.

\bibitem[Kim et~al., 2022]{kim2022occluded}
Kim, M., Cho, M., Lee, H., Cho, S., \& Lee, S. (2022).
\newblock Occluded person re-identification via relational adaptive feature correction learning.
\newblock In {\em ICASSP 2022-2022 IEEE International Conference on Acoustics, Speech and Signal Processing (ICASSP)}  (pp.\ 2719--2723).: IEEE.

\bibitem[Kiran et~al., 2021]{kiran2021holistic}
Kiran, M., Praveen, R.~G., Nguyen-Meidine, L.~T., Belharbi, S., Blais-Morin, L.-A., \& Granger, E. (2021).
\newblock Holistic guidance for occluded person re-identification.
\newblock {\em arXiv preprint arXiv:2104.06524}.

\bibitem[Lavi et~al., 2020]{lavi2020survey}
Lavi, B., Ullah, I., Fatan, M., \& Rocha, A. (2020).
\newblock Survey on reliable deep learning-based person re-identification models: Are we there yet?
\newblock {\em arXiv preprint arXiv:2005.00355}.

\bibitem[Leng et~al., 2019]{leng2019survey}
Leng, Q., Ye, M., \& Tian, Q. (2019).
\newblock A survey of open-world person re-identification.
\newblock {\em IEEE Transactions on Circuits and Systems for Video Technology}, 30(4), 1092--1108.

\bibitem[Li et~al., 2014]{li2014deepreid}
Li, W., Zhao, R., Xiao, T., \& Wang, X. (2014).
\newblock Deepreid: Deep filter pairing neural network for person re-identification.
\newblock In {\em Proceedings of the IEEE conference on computer vision and pattern recognition}  (pp.\ 152--159).

\bibitem[Li et~al., 2018]{li2018harmonious}
Li, W., Zhu, X., \& Gong, S. (2018).
\newblock Harmonious attention network for person re-identification.
\newblock In {\em Proceedings of the IEEE conference on computer vision and pattern recognition}  (pp.\ 2285--2294).

\bibitem[Li et~al., 2020]{li2020effective}
Li, Y., Jiang, X., \& Hwang, J.-N. (2020).
\newblock Effective person re-identification by self-attention model guided feature learning.
\newblock {\em Knowledge-Based Systems}, 187, 104832.

\bibitem[Li et~al., 2021]{li2021person}
Li, Y., Liu, L., Zhu, L., \& Zhang, H. (2021).
\newblock Person re-identification based on multi-scale feature learning.
\newblock {\em Knowledge-Based Systems}, 228, 107281.

\bibitem[Li et~al., 2019]{li2019recover}
Li, Y.-J., Chen, Y.-C., Lin, Y.-Y., Du, X., \& Wang, Y.-C.~F. (2019).
\newblock Recover and identify: A generative dual model for cross-resolution person re-identification.
\newblock In {\em Proceedings of the IEEE/CVF international conference on computer vision}  (pp.\ 8090--8099).

\bibitem[Liang et~al., 2018]{liang2018look}
Liang, X., Gong, K., Shen, X., \& Lin, L. (2018).
\newblock Look into person: Joint body parsing \& pose estimation network and a new benchmark.
\newblock {\em IEEE transactions on pattern analysis and machine intelligence}, 41(4), 871--885.

\bibitem[Lin \& Wang, 2021]{lin2021self}
Lin, C.-S. \& Wang, Y.-C.~F. (2021).
\newblock Self-supervised bodymap-to-appearance co-attention for partial person re-identification.
\newblock In {\em 2021 IEEE International Conference on Image Processing (ICIP)}  (pp.\ 2299--2303).: IEEE.

\bibitem[Lin et~al., 2014]{lin2014microsoft}
Lin, T.-Y., Maire, M., Belongie, S., Hays, J., Perona, P., Ramanan, D., Doll{\'a}r, P., \& Zitnick, C.~L. (2014).
\newblock Microsoft coco: Common objects in context.
\newblock In {\em Computer Vision--ECCV 2014: 13th European Conference, Zurich, Switzerland, September 6-12, 2014, Proceedings, Part V 13}  (pp.\ 740--755).: Springer.

\bibitem[Liu et~al., 2017]{liu2017neural}
Liu, H., Feng, J., Jie, Z., Jayashree, K., Zhao, B., Qi, M., Jiang, J., \& Yan, S. (2017).
\newblock Neural person search machines.
\newblock In {\em Proceedings of the IEEE International Conference on Computer Vision}  (pp.\ 493--501).

\bibitem[Liu et~al., 2019]{liu2019deep}
Liu, J., Zha, Z.-J., Hong, R., Wang, M., \& Zhang, Y. (2019).
\newblock Deep adversarial graph attention convolution network for text-based person search.
\newblock In {\em Proceedings of the 27th ACM International Conference on Multimedia}  (pp.\ 665--673).

\bibitem[Liu et~al., 2022]{liu2022dual}
Liu, Q., Teng, Q., Chen, H., Li, B., \& Qing, L. (2022).
\newblock Dual adaptive alignment and partitioning network for visible and infrared cross-modality person re-identification.
\newblock {\em Applied Intelligence}, 52(1), 547--563.

\bibitem[Liu et~al., 2021]{liu2021swin}
Liu, Z., Lin, Y., Cao, Y., Hu, H., Wei, Y., Zhang, Z., Lin, S., \& Guo, B. (2021).
\newblock Swin transformer: Hierarchical vision transformer using shifted windows.
\newblock In {\em Proceedings of the IEEE/CVF international conference on computer vision}  (pp.\ 10012--10022).

\bibitem[Loper et~al., 2015]{loper2015smpl}
Loper, M., Mahmood, N., Romero, J., Pons-Moll, G., \& Black, M.~J. (2015).
\newblock Smpl: A skinned multi-person linear model.
\newblock {\em ACM transactions on graphics (TOG)}, 34(6), 1--16.

\bibitem[Luo et~al., 2019]{luo2019strong}
Luo, H., Jiang, W., Gu, Y., Liu, F., Liao, X., Lai, S., \& Gu, J. (2019).
\newblock A strong baseline and batch normneuralization neck for deep person reidentification.
\newblock {\em arXiv preprint arXiv:1906.08332}.

\bibitem[Ma et~al., 2021]{ma2021pose}
Ma, Z., Zhao, Y., \& Li, J. (2021).
\newblock Pose-guided inter-and intra-part relational transformer for occluded person re-identification.
\newblock In {\em Proceedings of the 29th ACM International Conference on Multimedia}  (pp.\ 1487--1496).

\bibitem[Mao et~al., 2019]{mao2019resolution}
Mao, S., Zhang, S., \& Yang, M. (2019).
\newblock Resolution-invariant person re-identification.
\newblock {\em arXiv preprint arXiv:1906.09748}.

\bibitem[Martinel et~al., 2016]{martinel2016temporal}
Martinel, N., Das, A., Micheloni, C., \& Roy-Chowdhury, A.~K. (2016).
\newblock Temporal model adaptation for person re-identification.
\newblock In {\em Computer Vision--ECCV 2016: 14th European Conference, Amsterdam, The Netherlands, October 11--14, 2016, Proceedings, Part IV 14}  (pp.\ 858--877).: Springer.

\bibitem[Miao et~al., 2019]{miao2019pose}
Miao, J., Wu, Y., Liu, P., Ding, Y., \& Yang, Y. (2019).
\newblock Pose-guided feature alignment for occluded person re-identification.
\newblock In {\em Proceedings of the IEEE/CVF international conference on computer vision}  (pp.\ 542--551).

\bibitem[Miao et~al., 2021]{miao2021identifying}
Miao, J., Wu, Y., \& Yang, Y. (2021).
\newblock Identifying visible parts via pose estimation for occluded person re-identification.
\newblock {\em IEEE transactions on neural networks and learning systems}, 33(9), 4624--4634.

\bibitem[Ming et~al., 2022]{ming2022deep}
Ming, Z., Zhu, M., Wang, X., Zhu, J., Cheng, J., Gao, C., Yang, Y., \& Wei, X. (2022).
\newblock Deep learning-based person re-identification methods: A survey and outlook of recent works.
\newblock {\em Image and Vision Computing}, 119, 104394.

\bibitem[Nagaraju et~al., 2016]{nagaraju2016hierarchical}
Nagaraju, G., Raju, G. S.~R., Ko, Y.~H., \& Yu, J.~S. (2016).
\newblock Hierarchical ni--co layered double hydroxide nanosheets entrapped on conductive textile fibers: a cost-effective and flexible electrode for high-performance pseudocapacitors.
\newblock {\em Nanoscale}, 8(2), 812--825.

\bibitem[Nguyen et~al., 2017]{nguyen2017person}
Nguyen, D.~T., Hong, H.~G., Kim, K.~W., \& Park, K.~R. (2017).
\newblock Person recognition system based on a combination of body images from visible light and thermal cameras.
\newblock {\em Sensors}, 17(3), 605.

\bibitem[Ning et~al., 2021]{ning2021jwsaa}
Ning, X., Gong, K., Li, W., \& Zhang, L. (2021).
\newblock Jwsaa: joint weak saliency and attention aware for person re-identification.
\newblock {\em Neurocomputing}, 453, 801--811.

\bibitem[Ning et~al., 2020a]{ning2020feature}
Ning, X., Gong, K., Li, W., Zhang, L., Bai, X., \& Tian, S. (2020a).
\newblock Feature refinement and filter network for person re-identification.
\newblock {\em IEEE Transactions on Circuits and Systems for Video Technology}, 31(9), 3391--3402.

\bibitem[Ning et~al., 2020b]{ning2020multi}
Ning, X., Nan, F., Xu, S., Yu, L., \& Zhang, L. (2020b).
\newblock Multi-view frontal face image generation: a survey.
\newblock {\em Concurrency and Computation: Practice and Experience}, (pp.\ e6147).

\bibitem[Park et~al., 2021]{park2021learning}
Park, H., Lee, S., Lee, J., \& Ham, B. (2021).
\newblock Learning by aligning: Visible-infrared person re-identification using cross-modal correspondences.
\newblock In {\em Proceedings of the IEEE/CVF international conference on computer vision}  (pp.\ 12046--12055).

\bibitem[Peng et~al., 2022]{peng2022deep}
Peng, Y., Hou, S., Cao, C., Liu, X., Huang, Y., \& He, Z. (2022).
\newblock Deep learning-based occluded person re-identification: A survey.
\newblock {\em arXiv preprint arXiv:2207.14452}.

\bibitem[Qi et~al., 2017]{qi2017pointnet}
Qi, C.~R., Su, H., Mo, K., \& Guibas, L.~J. (2017).
\newblock Pointnet: Deep learning on point sets for 3d classification and segmentation.
\newblock In {\em Proceedings of the IEEE conference on computer vision and pattern recognition}  (pp.\ 652--660).

\bibitem[Quispe \& Pedrini, 2019]{quispe2019improved}
Quispe, R. \& Pedrini, H. (2019).
\newblock Improved person re-identification based on saliency and semantic parsing with deep neural network models.
\newblock {\em Image and Vision Computing}, 92, 103809.

\bibitem[Ren et~al., 2020]{ren2020semantic}
Ren, X., Zhang, D., \& Bao, X. (2020).
\newblock Semantic-guided shared feature alignment for occluded person re-identification.
\newblock In {\em Asian Conference on Machine Learning}  (pp.\ 17--32).: PMLR.

\bibitem[Ristani et~al., 2016]{ristani2016performance}
Ristani, E., Solera, F., Zou, R., Cucchiara, R., \& Tomasi, C. (2016).
\newblock Performance measures and a data set for multi-target, multi-camera tracking.
\newblock In {\em Computer Vision--ECCV 2016 Workshops: Amsterdam, The Netherlands, October 8-10 and 15-16, 2016, Proceedings, Part II}  (pp.\ 17--35).: Springer.

\bibitem[Sarfraz et~al., 2018]{sarfraz2018pose}
Sarfraz, M.~S., Schumann, A., Eberle, A., \& Stiefelhagen, R. (2018).
\newblock A pose-sensitive embedding for person re-identification with expanded cross neighborhood re-ranking.
\newblock In {\em Proceedings of the IEEE conference on computer vision and pattern recognition}  (pp.\ 420--429).

\bibitem[Sekhar et~al., 2017]{sekhar2017conductive}
Sekhar, S.~C., Nagaraju, G., \& Yu, J.~S. (2017).
\newblock Conductive silver nanowires-fenced carbon cloth fibers-supported layered double hydroxide nanosheets as a flexible and binder-free electrode for high-performance asymmetric supercapacitors.
\newblock {\em Nano Energy}, 36, 58--67.

\bibitem[Shamshad et~al., 2023]{shamshad2023transformers}
Shamshad, F., Khan, S., Zamir, S.~W., Khan, M.~H., Hayat, M., Khan, F.~S., \& Fu, H. (2023).
\newblock Transformers in medical imaging: A survey.
\newblock {\em Medical Image Analysis}, (pp.\ 102802).

\bibitem[Su et~al., 2017]{su2017pose}
Su, C., Li, J., Zhang, S., Xing, J., Gao, W., \& Tian, Q. (2017).
\newblock Pose-driven deep convolutional model for person re-identification.
\newblock In {\em Proceedings of the IEEE international conference on computer vision}  (pp.\ 3960--3969).

\bibitem[Sun et~al., 2019a]{sun2019deep}
Sun, K., Xiao, B., Liu, D., \& Wang, J. (2019a).
\newblock Deep high-resolution representation learning for human pose estimation.
\newblock In {\em Proceedings of the IEEE/CVF conference on computer vision and pattern recognition}  (pp.\ 5693--5703).

\bibitem[Sun \& Zheng, 2019]{sun2019dissecting}
Sun, X. \& Zheng, L. (2019).
\newblock Dissecting person re-identification from the viewpoint of viewpoint.
\newblock In {\em Proceedings of the IEEE/CVF conference on computer vision and pattern recognition}  (pp.\ 608--617).

\bibitem[Sun et~al., 2019b]{sun2019perceive}
Sun, Y., Xu, Q., Li, Y., Zhang, C., Li, Y., Wang, S., \& Sun, J. (2019b).
\newblock Perceive where to focus: Learning visibility-aware part-level features for partial person re-identification.
\newblock In {\em Proceedings of the IEEE/CVF conference on computer vision and pattern recognition}  (pp.\ 393--402).

\bibitem[Sun et~al., 2018]{sun2018beyond}
Sun, Y., Zheng, L., Yang, Y., Tian, Q., \& Wang, S. (2018).
\newblock Beyond part models: Person retrieval with refined part pooling (and a strong convolutional baseline).
\newblock In {\em Proceedings of the European conference on computer vision (ECCV)}  (pp.\ 480--496).

\bibitem[Szegedy et~al., 2016]{szegedy2016rethinking}
Szegedy, C., Vanhoucke, V., Ioffe, S., Shlens, J., \& Wojna, Z. (2016).
\newblock Rethinking the inception architecture for computer vision.
\newblock In {\em Proceedings of the IEEE conference on computer vision and pattern recognition}  (pp.\ 2818--2826).

\bibitem[Tan et~al., 2021]{tan2021incomplete}
Tan, H., Liu, X., Bian, Y., Wang, H., \& Yin, B. (2021).
\newblock Incomplete descriptor mining with elastic loss for person re-identification.
\newblock {\em IEEE Transactions on Circuits and Systems for Video Technology}, 32(1), 160--171.

\bibitem[Tan et~al., 2022a]{tan2022mhsa}
Tan, H., Liu, X., Yin, B., \& Li, X. (2022a).
\newblock Mhsa-net: Multihead self-attention network for occluded person re-identification.
\newblock {\em IEEE Transactions on Neural Networks and Learning Systems}.

\bibitem[Tan et~al., 2022b]{tan2022dynamic}
Tan, L., Dai, P., Ji, R., \& Wu, Y. (2022b).
\newblock Dynamic prototype mask for occluded person re-identification.
\newblock In {\em Proceedings of the 30th ACM International Conference on Multimedia}  (pp.\ 531--540).

\bibitem[Tirkolaee et~al., 2020]{tirkolaee2020fuzzy}
Tirkolaee, E.~B., Goli, A., \& Weber, G.-W. (2020).
\newblock Fuzzy mathematical programming and self-adaptive artificial fish swarm algorithm for just-in-time energy-aware flow shop scheduling problem with outsourcing option.
\newblock {\em IEEE transactions on fuzzy systems}, 28(11), 2772--2783.

\bibitem[Tutsoy et~al., 2018]{tutsoy2018design}
Tutsoy, O., Barkana, D.~E., \& Tugal, H. (2018).
\newblock Design of a completely model free adaptive control in the presence of parametric, non-parametric uncertainties and random control signal delay.
\newblock {\em ISA transactions}, 76, 67--77.

\bibitem[Tutsoy \& Tanrikulu, 2022]{tutsoy2022priority}
Tutsoy, O. \& Tanrikulu, M.~Y. (2022).
\newblock Priority and age specific vaccination algorithm for the pandemic diseases: a comprehensive parametric prediction model.
\newblock {\em BMC Medical Informatics and Decision Making}, 22(1), 4.

\bibitem[Wang et~al., 2022a]{wang2022learning}
Wang, C., Ning, X., Sun, L., Zhang, L., Li, W., \& Bai, X. (2022a).
\newblock Learning discriminative features by covering local geometric space for point cloud analysis.
\newblock {\em IEEE Transactions on Geoscience and Remote Sensing}, 60, 1--15.

\bibitem[Wang et~al., 2021a]{wang2021brief}
Wang, C., Wang, C., Li, W., \& Wang, H. (2021a).
\newblock A brief survey on rgb-d semantic segmentation using deep learning.
\newblock {\em Displays}, 70, 102080.

\bibitem[Wang et~al., 2020a]{wang2020high}
Wang, G., Yang, S., Liu, H., Wang, Z., Yang, Y., Wang, S., Yu, G., Zhou, E., \& Sun, J. (2020a).
\newblock High-order information matters: Learning relation and topology for occluded person re-identification.
\newblock In {\em Proceedings of the IEEE/CVF conference on computer vision and pattern recognition}  (pp.\ 6449--6458).

\bibitem[Wang et~al., 2019a]{wang2019rgb}
Wang, G., Zhang, T., Cheng, J., Liu, S., Yang, Y., \& Hou, Z. (2019a).
\newblock Rgb-infrared cross-modality person re-identification via joint pixel and feature alignment.
\newblock In {\em Proceedings of the IEEE/CVF International Conference on Computer Vision}  (pp.\ 3623--3632).

\bibitem[Wang et~al., 2021b]{wang2021deep}
Wang, J., Tan, S., Zhen, X., Xu, S., Zheng, F., He, Z., \& Shao, L. (2021b).
\newblock Deep 3d human pose estimation: A review.
\newblock {\em Computer Vision and Image Understanding}, 210, 103225.

\bibitem[Wang et~al., 2022b]{wang2022quality}
Wang, P., Ding, C., Shao, Z., Hong, Z., Zhang, S., \& Tao, D. (2022b).
\newblock Quality-aware part models for occluded person re-identification.
\newblock {\em IEEE Transactions on Multimedia}.

\bibitem[Wang et~al., 2022c]{wang2022swin}
Wang, Q., Huang, H., Zhong, Y., Min, W., Han, Q., Xu, D., \& Xu, C. (2022c).
\newblock Swin transformer based on two-fold loss and background adaptation re-ranking for person re-identification.
\newblock {\em Electronics}, 11(13), 1941.

\bibitem[Wang et~al., 2020b]{wang2020deep}
Wang, Q., Qi, M., Jin, K., \& Jiang, J. (2020b).
\newblock Deep-shallow occlusion parallelism network for person re-identification.
\newblock In {\em Journal of Physics: Conference Series}, volume 1518  (pp.\ 012026).: IOP Publishing.

\bibitem[Wang et~al., 2022d]{wang2022pose}
Wang, T., Liu, H., Song, P., Guo, T., \& Shi, W. (2022d).
\newblock Pose-guided feature disentangling for occluded person re-identification based on transformer.
\newblock In {\em Proceedings of the AAAI Conference on Artificial Intelligence}, volume~36  (pp.\ 2540--2549).

\bibitem[Wang et~al., 2022e]{wang2022cross}
Wang, X., Li, C., \& Ma, X. (2022e).
\newblock Cross-modal local shortest path and global enhancement for visible-thermal person re-identification.
\newblock {\em arXiv preprint arXiv:2206.04401}.

\bibitem[Wang et~al., 2022f]{wang2022cloning}
Wang, Y., Liang, X., \& Liao, S. (2022f).
\newblock Cloning outfits from real-world images to 3d characters for generalizable person re-identification.
\newblock In {\em Proceedings of the IEEE/CVF Conference on Computer Vision and Pattern Recognition}  (pp.\ 4900--4909).

\bibitem[Wang et~al., 2019b]{wang2019beyond}
Wang, Z., Wang, Z., Wu, Y., Wang, J., \& Satoh, S. (2019b).
\newblock Beyond intra-modality discrepancy: A comprehensive survey of heterogeneous person re-identification.
\newblock {\em arXiv preprint arXiv:1905.10048}, 4.

\bibitem[Wang et~al., 2019c]{wang2019learning}
Wang, Z., Wang, Z., Zheng, Y., Chuang, Y.-Y., \& Satoh, S. (2019c).
\newblock Learning to reduce dual-level discrepancy for infrared-visible person re-identification.
\newblock In {\em Proceedings of the IEEE/CVF conference on computer vision and pattern recognition}  (pp.\ 618--626).

\bibitem[Wang et~al., 2022g]{wang2022feature}
Wang, Z., Zhu, F., Tang, S., Zhao, R., He, L., \& Song, J. (2022g).
\newblock Feature erasing and diffusion network for occluded person re-identification.
\newblock In {\em Proceedings of the IEEE/CVF Conference on Computer Vision and Pattern Recognition}  (pp.\ 4754--4763).

\bibitem[Wen et~al., 2022]{wen2022cross}
Wen, X., Feng, X., Li, P., \& Chen, W. (2022).
\newblock Cross-modality collaborative learning identified pedestrian.
\newblock {\em The Visual Computer}, (pp.\ 1--16).

\bibitem[Wu et~al., 2017a]{wu2017robust}
Wu, A., Zheng, W.-S., \& Lai, J.-H. (2017a).
\newblock Robust depth-based person re-identification.
\newblock {\em IEEE Transactions on Image Processing}, 26(6), 2588--2603.

\bibitem[Wu et~al., 2017b]{wu2017rgb}
Wu, A., Zheng, W.-S., Yu, H.-X., Gong, S., \& Lai, J. (2017b).
\newblock Rgb-infrared cross-modality person re-identification.
\newblock In {\em Proceedings of the IEEE international conference on computer vision}  (pp.\ 5380--5389).

\bibitem[Wu et~al., 2018]{wu2018exploit}
Wu, Y., Lin, Y., Dong, X., Yan, Y., Ouyang, W., \& Yang, Y. (2018).
\newblock Exploit the unknown gradually: One-shot video-based person re-identification by stepwise learning.
\newblock In {\em Proceedings of the IEEE conference on computer vision and pattern recognition}  (pp.\ 5177--5186).

\bibitem[Xie et~al., 2020]{xie2020linking}
Xie, Y., Tian, J., \& Zhu, X.~X. (2020).
\newblock Linking points with labels in 3d: A review of point cloud semantic segmentation.
\newblock {\em IEEE Geoscience and remote sensing magazine}, 8(4), 38--59.

\bibitem[Xu et~al., 2022]{xu2022learning}
Xu, B., He, L., Liang, J., \& Sun, Z. (2022).
\newblock Learning feature recovery transformer for occluded person re-identification.
\newblock {\em IEEE Transactions on Image Processing}, 31, 4651--4662.

\bibitem[Xu et~al., 2018]{xu2018attention}
Xu, J., Zhao, R., Zhu, F., Wang, H., \& Ouyang, W. (2018).
\newblock Attention-aware compositional network for person re-identification.
\newblock In {\em Proceedings of the IEEE conference on computer vision and pattern recognition}  (pp.\ 2119--2128).

\bibitem[Xu et~al., 2021]{xu2021dual}
Xu, Y., Zhao, L., \& Qin, F. (2021).
\newblock Dual attention-based method for occluded person re-identification.
\newblock {\em Knowledge-Based Systems}, 212, 106554.

\bibitem[Yaghoubi et~al., 2021]{yaghoubi2021sss}
Yaghoubi, E., Kumar, A., \& Proen{\c{c}}a, H. (2021).
\newblock Sss-pr: A short survey of surveys in person re-identification.
\newblock {\em Pattern Recognition Letters}, 143, 50--57.

\bibitem[Yan et~al., 2021]{yan2021occluded}
Yan, C., Pang, G., Jiao, J., Bai, X., Feng, X., \& Shen, C. (2021).
\newblock Occluded person re-identification with single-scale global representations.
\newblock In {\em Proceedings of the IEEE/CVF International Conference on Computer Vision}  (pp.\ 11875--11884).

\bibitem[Yang et~al., 2022]{yang2022pafm}
Yang, J., Zhang, C., Tang, Y., \& Li, Z. (2022).
\newblock Pafm: pose-drive attention fusion mechanism for occluded person re-identification.
\newblock {\em Neural Computing and Applications}, 34(10), 8241--8252.

\bibitem[Yang et~al., 2021]{yang2021learning}
Yang, J., Zhang, J., Yu, F., Jiang, X., Zhang, M., Sun, X., Chen, Y.-C., \& Zheng, W.-S. (2021).
\newblock Learning to know where to see: A visibility-aware approach for occluded person re-identification.
\newblock In {\em Proceedings of the IEEE/CVF international conference on computer vision}  (pp.\ 11885--11894).

\bibitem[Yang et~al., 2020]{Yang2020}
Yang, W., Yan, Y., Chen, S., Zhang, X., \& Wang, H. (2020).
\newblock Multi-scale generative adversarial network for person reidentification under occlusion.
\newblock {\em Journal of Software}, 31(7), 1943--1958.

\bibitem[Ye et~al., 2021a]{ye2021dynamic}
Ye, M., Chen, C., Shen, J., \& Shao, L. (2021a).
\newblock Dynamic tri-level relation mining with attentive graph for visible infrared re-identification.
\newblock {\em IEEE Transactions on Information Forensics and Security}, 17, 386--398.

\bibitem[Ye et~al., 2020]{ye2020dynamic}
Ye, M., Shen, J., J.~Crandall, D., Shao, L., \& Luo, J. (2020).
\newblock Dynamic dual-attentive aggregation learning for visible-infrared person re-identification.
\newblock In {\em Computer Vision--ECCV 2020: 16th European Conference, Glasgow, UK, August 23--28, 2020, Proceedings, Part XVII 16}  (pp.\ 229--247).: Springer.

\bibitem[Ye et~al., 2021b]{ye2021deep}
Ye, M., Shen, J., Lin, G., Xiang, T., Shao, L., \& Hoi, S.~C. (2021b).
\newblock Deep learning for person re-identification: A survey and outlook.
\newblock {\em IEEE transactions on pattern analysis and machine intelligence}, 44(6), 2872--2893.

\bibitem[Zhai et~al., 2021]{zhai2021pgmanet}
Zhai, Y., Han, X., Ma, W., Gou, X., \& Xiao, G. (2021).
\newblock Pgmanet: Pose-guided mixed attention network for occluded person re-identification.
\newblock In {\em 2021 International Joint Conference on Neural Networks (IJCNN)}  (pp.\ 1--8).: IEEE.

\bibitem[Zhang et~al., 2021]{zhang2021multi}
Zhang, C., Liu, H., Guo, W., \& Ye, M. (2021).
\newblock Multi-scale cascading network with compact feature learning for rgb-infrared person re-identification.
\newblock In {\em 2020 25th International Conference on Pattern Recognition (ICPR)}  (pp.\ 8679--8686).: IEEE.

\bibitem[Zhang et~al., 2022a]{zhang2022fine}
Zhang, G., Chen, C., Chen, Y., Zhang, H., \& Zheng, Y. (2022a).
\newblock Fine-grained-based multi-feature fusion for occluded person re-identification.
\newblock {\em Journal of Visual Communication and Image Representation}, 87, 103581.

\bibitem[Zhang et~al., 2022b]{zhang2022hybrid}
Zhang, L., Guo, H., Zhu, K., Qiao, H., Huang, G., Zhang, S., Zhang, H., Sun, J., \& Wang, J. (2022b).
\newblock Hybrid modality metric learning for visible-infrared person re-identification.
\newblock {\em ACM Transactions on Multimedia Computing, Communications, and Applications (TOMM)}, 18(1s), 1--15.

\bibitem[Zhang et~al., 2022c]{zhang2022modeling}
Zhang, Q., Dang, K., Lai, J.-H., Feng, Z., \& Xie, X. (2022c).
\newblock Modeling 3d layout for group re-identification.
\newblock In {\em Proceedings of the IEEE/CVF Conference on Computer Vision and Pattern Recognition}  (pp.\ 7512--7520).

\bibitem[Zhang et~al., 2020]{zhang2020semantic}
Zhang, X., Yan, Y., Xue, J.-H., Hua, Y., \& Wang, H. (2020).
\newblock Semantic-aware occlusion-robust network for occluded person re-identification.
\newblock {\em IEEE Transactions on Circuits and Systems for Video Technology}, 31(7), 2764--2778.

\bibitem[Zhang \& Lu, 2018]{zhang2018deep}
Zhang, Y. \& Lu, H. (2018).
\newblock Deep cross-modal projection learning for image-text matching.
\newblock In {\em Proceedings of the European conference on computer vision (ECCV)}  (pp.\ 686--701).

\bibitem[Zhang et~al., 2019]{zhang2019densely}
Zhang, Z., Lan, C., Zeng, W., \& Chen, Z. (2019).
\newblock Densely semantically aligned person re-identification.
\newblock In {\em Proceedings of the IEEE/CVF conference on computer vision and pattern recognition}  (pp.\ 667--676).

\bibitem[Zhao et~al., 2021]{zhao2021incremental}
Zhao, C., Lv, X., Dou, S., Zhang, S., Wu, J., \& Wang, L. (2021).
\newblock Incremental generative occlusion adversarial suppression network for person reid.
\newblock {\em IEEE Transactions on Image Processing}, 30, 4212--4224.

\bibitem[Zhao et~al., 2013]{zhao2013unsupervised}
Zhao, R., Ouyang, W., \& Wang, X. (2013).
\newblock Unsupervised salience learning for person re-identification.
\newblock In {\em Proceedings of the IEEE conference on computer vision and pattern recognition}  (pp.\ 3586--3593).

\bibitem[Zhao et~al., 2020]{zhao2020not}
Zhao, S., Gao, C., Zhang, J., Cheng, H., Han, C., Jiang, X., Guo, X., Zheng, W.-S., Sang, N., \& Sun, X. (2020).
\newblock Do not disturb me: Person re-identification under the interference of other pedestrians.
\newblock In {\em Computer Vision--ECCV 2020: 16th European Conference, Glasgow, UK, August 23--28, 2020, Proceedings, Part VI 16}  (pp.\ 647--663).: Springer.

\bibitem[Zhao et~al., 2022]{zhao2022short}
Zhao, Y., Zhu, S., Wang, D., \& Liang, Z. (2022).
\newblock Short range correlation transformer for occluded person re-identification.
\newblock {\em Neural computing and applications}, 34(20), 17633--17645.

\bibitem[Zheng et~al., 2021]{zheng2021pose}
Zheng, K., Lan, C., Zeng, W., Liu, J., Zhang, Z., \& Zha, Z.-J. (2021).
\newblock Pose-guided feature learning with knowledge distillation for occluded person re-identification.
\newblock In {\em Proceedings of the 29th ACM International Conference on Multimedia}  (pp.\ 4537--4545).

\bibitem[Zheng et~al., 2015a]{zheng2015scalable}
Zheng, L., Shen, L., Tian, L., Wang, S., Wang, J., \& Tian, Q. (2015a).
\newblock Scalable person re-identification: A benchmark.
\newblock In {\em Proceedings of the IEEE international conference on computer vision}  (pp.\ 1116--1124).

\bibitem[Zheng et~al., 2016]{zheng2016person}
Zheng, L., Yang, Y., \& Hauptmann, A.~G. (2016).
\newblock Person re-identification: Past, present and future.
\newblock {\em arXiv preprint arXiv:1610.02984}.

\bibitem[Zheng et~al., 2019]{zheng2019re}
Zheng, M., Karanam, S., Wu, Z., \& Radke, R.~J. (2019).
\newblock Re-identification with consistent attentive siamese networks.
\newblock In {\em Proceedings of the IEEE/CVF conference on computer vision and pattern recognition}  (pp.\ 5735--5744).

\bibitem[Zheng et~al., 2011]{zheng2011person}
Zheng, W.-S., Gong, S., \& Xiang, T. (2011).
\newblock Person re-identification by probabilistic relative distance comparison.
\newblock In {\em CVPR 2011}  (pp.\ 649--656).: IEEE.

\bibitem[Zheng et~al., 2015b]{zheng2015partial}
Zheng, W.-S., Li, X., Xiang, T., Liao, S., Lai, J., \& Gong, S. (2015b).
\newblock Partial person re-identification.
\newblock In {\em Proceedings of the IEEE international conference on computer vision}  (pp.\ 4678--4686).

\bibitem[Zheng et~al., 2022]{zheng2022parameter}
Zheng, Z., Wang, X., Zheng, N., \& Yang, Y. (2022).
\newblock Parameter-efficient person re-identification in the 3d space.
\newblock {\em IEEE Transactions on Neural Networks and Learning Systems}.

\bibitem[Zheng et~al., 2017]{zheng2017unlabeled}
Zheng, Z., Zheng, L., \& Yang, Y. (2017).
\newblock Unlabeled samples generated by gan improve the person re-identification baseline in vitro.
\newblock In {\em Proceedings of the IEEE international conference on computer vision}  (pp.\ 3754--3762).

\bibitem[Zhong et~al., 2020a]{zhong2020robust}
Zhong, Y., Wang, X., \& Zhang, S. (2020a).
\newblock Robust partial matching for person search in the wild.
\newblock In {\em Proceedings of the IEEE/CVF conference on computer vision and pattern recognition}  (pp.\ 6827--6835).

\bibitem[Zhong et~al., 2020b]{zhong2020random}
Zhong, Z., Zheng, L., Kang, G., Li, S., \& Yang, Y. (2020b).
\newblock Random erasing data augmentation.
\newblock In {\em Proceedings of the AAAI conference on artificial intelligence}, volume~34  (pp.\ 13001--13008).

\bibitem[Zhou et~al., 2019]{zhou2019omni}
Zhou, K., Yang, Y., Cavallaro, A., \& Xiang, T. (2019).
\newblock Omni-scale feature learning for person re-identification.
\newblock In {\em Proceedings of the IEEE/CVF international conference on computer vision}  (pp.\ 3702--3712).

\bibitem[Zhou et~al., 2022]{zhou2022motion}
Zhou, M., Liu, H., Lv, Z., Hong, W., \& Chen, X. (2022).
\newblock Motion-aware transformer for occluded person re-identification.
\newblock {\em arXiv preprint arXiv:2202.04243}.

\bibitem[Zhou et~al., 2020a]{zhou2020fine}
Zhou, Q., Zhong, B., Lan, X., Sun, G., Zhang, Y., Zhang, B., \& Ji, R. (2020a).
\newblock Fine-grained spatial alignment model for person re-identification with focal triplet loss.
\newblock {\em IEEE Transactions on Image Processing}, 29, 7578--7589.

\bibitem[Zhou et~al., 2020b]{zhou2020depth}
Zhou, S., Wu, J., Zhang, F., \& Sehdev, P. (2020b).
\newblock Depth occlusion perception feature analysis for person re-identification.
\newblock {\em Pattern Recognition Letters}, 138, 617--623.

\bibitem[Zhu et~al., 2020]{zhu2020identity}
Zhu, K., Guo, H., Liu, Z., Tang, M., \& Wang, J. (2020).
\newblock Identity-guided human semantic parsing for person re-identification.
\newblock In {\em Computer Vision--ECCV 2020: 16th European Conference, Glasgow, UK, August 23--28, 2020, Proceedings, Part III 16}  (pp.\ 346--363).: Springer.

\bibitem[Zhu et~al., 2022]{zhu2022information}
Zhu, X., Zheng, M., Chen, X., Zhang, X., Yuan, C., \& Zhang, F. (2022).
\newblock Information disentanglement based cross-modal representation learning for visible-infrared person re-identification.
\newblock {\em Multimedia Tools and Applications}, (pp.\ 1--27).

\bibitem[Zhuo et~al., 2018]{zhuo2018occluded}
Zhuo, J., Chen, Z., Lai, J., \& Wang, G. (2018).
\newblock Occluded person re-identification.
\newblock In {\em 2018 IEEE International Conference on Multimedia and Expo (ICME)}  (pp.\ 1--6).: IEEE.

\bibitem[Zhuo et~al., 2019]{zhuo2019novel}
Zhuo, J., Lai, J., \& Chen, P. (2019).
\newblock A novel teacher-student learning framework for occluded person re-identification.
\newblock {\em arXiv preprint arXiv:1907.03253}.

\end{thebibliography}



\end{document}